\documentclass[11pt]{article}

\usepackage[preprint]{acl}

\usepackage{times}
\usepackage{latexsym}

\usepackage[T1]{fontenc}

\usepackage[utf8]{inputenc}

\usepackage{microtype}

\usepackage{inconsolata}

\usepackage{graphicx}
\usepackage{booktabs}
\usepackage{multirow}
\usepackage{listings}
\usepackage[most]{tcolorbox}
\usepackage[table]{xcolor}
\usepackage{makecell}
\usepackage{pifont}
\usepackage{enumitem}

\definecolor{bg_sft}{RGB}{255, 245, 245} 
\definecolor{bg_rl}{RGB}{245, 255, 245}  
\definecolor{border_gray}{RGB}{200, 200, 200}

\newcommand{\cmark}{\ding{51}} 
\newcommand{\xmark}{\ding{55}} 

\author{
 \textbf{Xuanyu Lei\textsuperscript{1,2}},
 \textbf{Yiqi Zhu\textsuperscript{2}},
 \textbf{Chenliang Li\textsuperscript{3}},
 \textbf{Kaiming Liu\textsuperscript{2}}, \\
 \textbf{Peng Li\textsuperscript{1,$\dagger$}}, 
 \textbf{Ming Yan\textsuperscript{3,$\dagger$}},
 \textbf{Jieping Ye\textsuperscript{3}},
 \textbf{Ya-Qin Zhang\textsuperscript{1}},
 \textbf{Yang Liu\textsuperscript{1,2,$\dagger$}} \\
 \textsuperscript{1}Institute for AI Industry Research (AIR), Tsinghua University, Beijing, China\\
 \textsuperscript{2}Dept. of Comp. Sci. \& Tech., Institute for AI, Tsinghua University, Beijing, China\\
 \textsuperscript{3}Institute of Intelligent Computing, Alibaba Group\\
 \texttt{leixy24@mails.tsinghua.edu.cn, lipeng@air.tsinghua.edu.cn} \\
 \texttt{ym119608@alibaba-inc.com, liuyang2011@tsinghua.edu.cn}
}

\title{\textsc{State2State}: Environment-Derived Mid-Training for LLM Agents}

\begin{document}
\maketitle
\begin{abstract}
Training LLM agents commonly relies on supervised fine-tuning from expert trajectories or online reinforcement learning over human-specified tasks with handcrafted verifiers. Though effective, both remain bottlenecked by externally specified tasks and supervision signals, limiting the scalability and diversity of agent training. 
We study an \textit{environment learning} paradigm in which agents acquire interaction and manipulation capabilities solely through environment interaction, without externally specified tasks.
We propose \textsc{State2State}, an environment-derived mid-training method that converts explored environment states into training objectives, challenging agents to reach a specified target state.
By deriving tasks from environment exploration and verifying success through rule-based state matching, \textsc{State2State} provides scalable and verifiable training objectives without expert supervision or manual task design.
Experiments on ALFWorld and ScienceWorld show that \textsc{State2State} improves agent performance as a standalone environment-learning stage in most settings. As initialization for downstream RL, it further improves final performance and learning efficiency, with promising evidence of cross-environment generalization.
\end{abstract}

\noindent\let\thefootnote\relax\footnotetext{$^\dagger$ Corresponding Authors.}
\noindent\let\thefootnote\relax\footnotetext{$^\ddagger$ Code will be released at \url{https://github.com/THUNLP-MT/State2State}.}

\section{Introduction}

Existing methods for training LLM agents mainly follow two paradigms: supervised fine-tuning~(SFT) from expert trajectories and reinforcement learning with verifiable rewards (RLVR). 
SFT trains agents to imitate trajectories collected from human experts or stronger teacher models~\cite{DBLP:conf/acl/ZengLLWLD024, DBLP:journals/corr/abs-2508-09123}, but such trajectories are costly to obtain, inherit the capability limits of their sources, and constrain learning to the coverage of a limited task distribution. 
RLVR reduces the reliance on expert demonstrations by allowing agents to learn through interaction~\cite{DBLP:journals/corr/abs-2504-20073, DBLP:journals/corr/abs-2505-10978, DBLP:journals/corr/abs-2503-09516}, but it shifts the bottleneck to task and reward construction: effective training still requires externally specified goals of suitable difficulty and manually designed verifiers for judging success.

\begin{figure}[t]
  \centering
  \includegraphics[width=\linewidth]{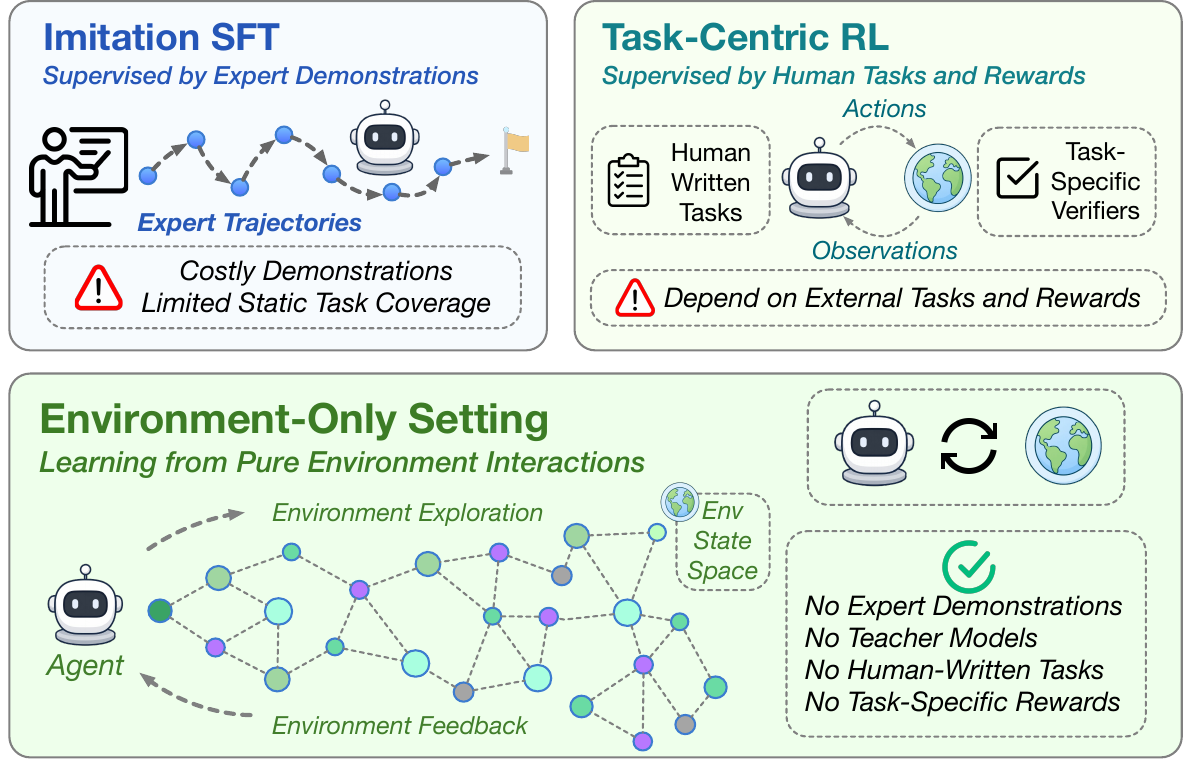}
  \vspace{-2mm}
  \caption{Comparison of supervision sources for LLM agent training. SFT relies on expert trajectories, and task-centric RL relies on human-specified tasks with task-specific verifiers. In contrast, the environment-only setting studied in this work derives learning signals purely from environment itself.}
  \label{fig:paradigm-comparison}
  \vspace{-5mm}
\end{figure}

Recent task generation methods~\cite{DBLP:conf/kdd/Hu0XSLLLR25, DBLP:journals/corr/abs-2506-10055} attempt to alleviate this bottleneck by expanding the supply of training tasks. For example, recent works~\cite{DBLP:journals/corr/abs-2511-10395, DBLP:journals/corr/abs-2509-15738} use stronger teacher models to collect interaction trajectories, synthesize task instructions, and construct domain-specific verifiers. While these methods improve the scalability of task acquisition, they still largely treat agent training as learning to solve externally formulated tasks. As a result, the diversity and depth of training objectives remain tied to human-task distributions, teacher-model capabilities, or domain-specific task construction pipelines.

Viewed through the lens of supervision signals, SFT relies on expert action trajectories, while RLVR relies on externally specified goals and verifiers. However, both paradigms remain centered on human-defined task distributions and external supervision sources. 
As illustrated in Figure~\ref{fig:paradigm-comparison}, this raises a natural question beyond action-level or goal-level supervision: \textit{Can agents learn useful environment perception and manipulation capabilities directly from environment-only supervision?} 
In this work, we study this question through an \emph{environment learning} paradigm. 
Environment learning considers a minimal training setting with only the agent and the environment, where supervision is derived from the environment itself, without external teachers or human-authored tasks. 
By first learning to understand and manipulate the environment, agents can acquire primitive environment skill priors that later serve as a foundation for RL on human-specified tasks, where these capabilities are aligned with human intent. 
This progression makes environment learning naturally suited for agent mid-training~\cite{DBLP:journals/corr/abs-2510-23081}, which aims to develop useful capabilities by intermediate training before optimizing specific downstream tasks.

\begin{figure*}[t]
    \centering
    \includegraphics[width=\linewidth]{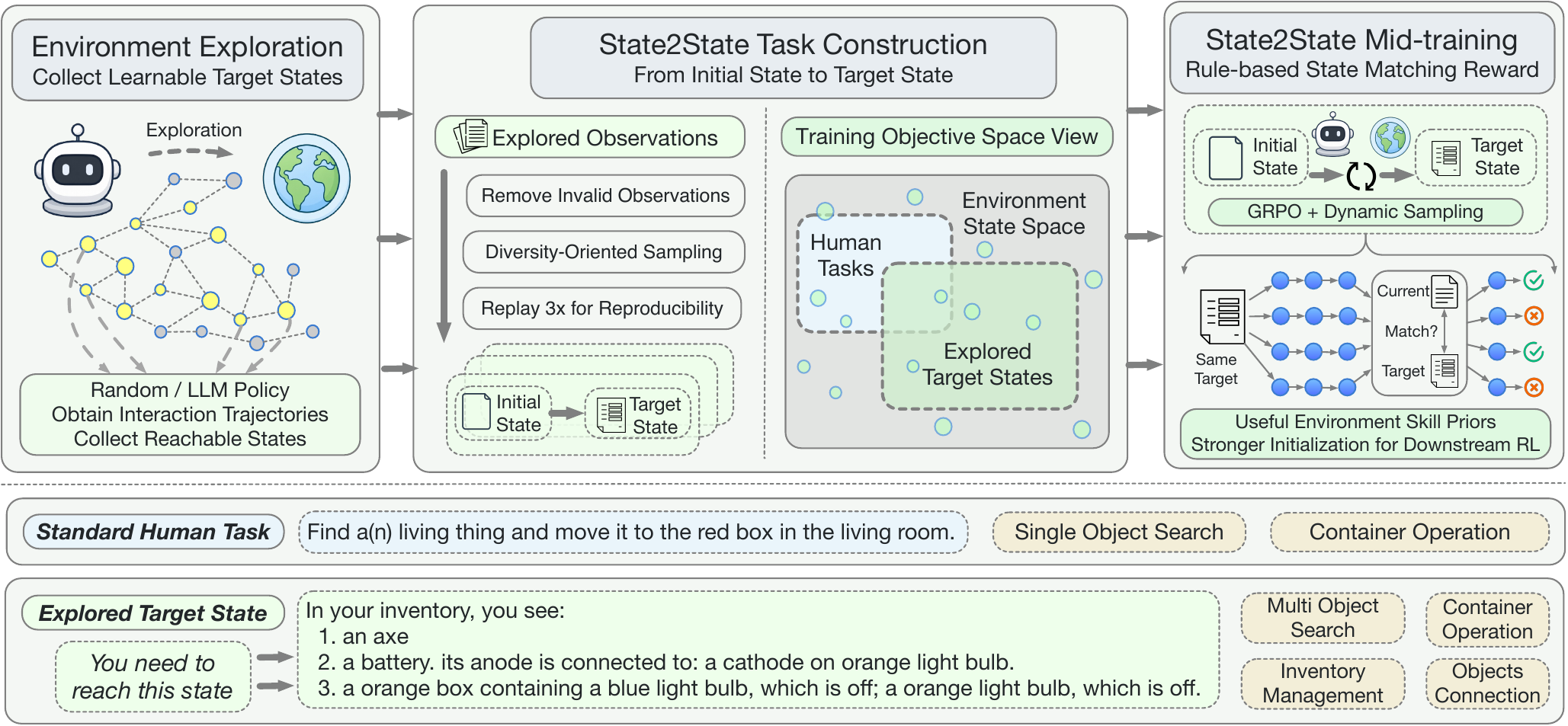}
    \vspace{-5mm}
    \caption{
    Overview of \textsc{State2State}, an environment-derived mid-training method.
    \textsc{State2State} converts explored environment states into reproducible training objectives and trains agents with rule-based state-matching rewards.
    By sampling targets from explorations, \textsc{State2State} aims to expose agents to a different and potentially broader set of environment states, not limited by human-task distributions.
    }
    \label{fig:framework}
    \vspace{-4mm}
\end{figure*}

Under this setting, we propose \textsc{State2State}, a simple instantiation of environment learning that turns reachable environment states into verifiable mid-training objectives. 
As shown in Figure~\ref{fig:framework}, \textsc{State2State} begins by exploring the environment with an exploration policy to collect reproducible reachable states. 
It then filters and samples learnable target states to improve diversity and reproducibility, and pairs each target with a corresponding initial configuration to construct a state-reaching task. 
In each task, given the initial observation and the target observation, the agent must reason and act through environment interactions to transform the current state into the target state. 
By sampling from explored states rather than predefined static tasks, \textsc{State2State} derives task diversity directly from the reachable environment state space, exposing agents to environment dynamics beyond human-designed task distributions. 
Task success can be verified by matching the resulting state against the target state, yielding rule-based rewards without task-specific test cases. 
We train agents on these objectives using GRPO-based reinforcement learning~\cite{DBLP:journals/corr/abs-2402-03300} with dynamic sampling~\cite{DBLP:journals/corr/abs-2503-14476}, which collects informative rollouts for effective and stable training. 
Overall, \textsc{State2State} turns environment exploration into verifiable RL objectives, allowing agents to learn from the environment itself without expert demonstrations, teacher models, or human-authored task instructions, while requiring only reproducible environment states and a rule-based state-matching verifier.
The resulting environment-aware policy serves as a stronger starting point for subsequent human-task RL, which further aligns the environment skill priors with human intent and achieves better task completion performance.


We evaluate \textsc{State2State} on ALFWorld~\cite{DBLP:conf/iclr/ShridharYCBTH21}, ScienceWorld~\cite{DBLP:conf/emnlp/WangJCA22}, covering household and scientific environments, and further conduct an extension to the more complex MobileWorld~\cite{DBLP:journals/corr/abs-2512-19432} GUI environment.
Across these environments, \textsc{State2State} improves agent performance before task-specific RL in most settings, showing that verifiable state-reaching objectives can induce useful environment understanding and manipulation capabilities. 
Serving as a stronger initialization for subsequent RL on human-specified tasks, \textsc{State2State} further improves final task performance and learning efficiency of downstream RL, demonstrating its effectiveness as an environment-centric mid-training stage. 
We also observe positive transfer from ScienceWorld to ALFWorld: models mid-trained with \textsc{State2State} generalize better than those trained on standard ScienceWorld tasks, suggesting that environment-derived objectives potentially support stronger cross-environment generalization.
Finally, \textsc{State2State} yields a standalone improvement on the GUI tasks of MobileWorld, providing initial evidence that the method can extend to more complex GUI interaction environments where collecting human-specified training tasks and reliable verifiers is costly and difficult.

In summary, our main contributions are:
\begin{itemize}[leftmargin=1.5em,itemsep=0pt,parsep=0.2em,topsep=0.1em,partopsep=0.0em]
\item We study \emph{environment learning} as a mid-training paradigm for LLM agents, where agents learn environment perception and manipulation capabilities from direct agent-environment interaction without external teachers or manually designed tasks.
\item We introduce \textsc{State2State}, which turns explored reachable states into verifiable state-reaching objectives and trains agents with state-matching rewards, enabling reinforcement learning from the environment itself.
\item We evaluate \textsc{State2State} on ALFWorld and ScienceWorld, demonstrating gains before task-specific RL in most settings, better initialization for subsequent human-task RL, and promising cross-environment generalization.
\end{itemize}

\section{Related Work}

\noindent\textbf{LLM Agent Training.}
Recent work has improved LLM agents through supervised fine-tuning and reinforcement learning on task-oriented interaction data. Early studies~\cite{DBLP:conf/acl/ZengLLWLD024, DBLP:journals/corr/abs-2310-05915} fine-tune LLMs on expert trajectories to elicit tool use and decision making. More recent efforts~\cite{DBLP:journals/corr/abs-2510-24701, DBLP:journals/corr/abs-2507-20534, DBLP:journals/corr/abs-2509-13310, DBLP:journals/corr/abs-2508-09123} further scale this paradigm by automatically constructing task instructions and corresponding trajectories. 
In parallel, agentic RL~\cite{DBLP:conf/iclr/QiLILSSYYY00D25, DBLP:journals/corr/abs-2504-20073} optimizes agents through online interactions and verifiable rewards.
Other studies further refine long-horizon behavior with more fine-grained optimization signals~\cite{DBLP:journals/corr/abs-2505-10978, DBLP:journals/corr/abs-2507-22844}.
Despite their effectiveness, these methods largely rely on predefined tasks, expert trajectories or manually designed rewards.

\noindent\textbf{Agentic Mid-Training.}
Agentic mid-training introduces intermediate objectives before final task training. Prior work~\cite{DBLP:journals/corr/abs-2510-23081} studies general mid-training for LLMs, and this idea has been extended to agent training by introducing intermediate objectives including world modeling~\cite{DBLP:journals/corr/abs-2604-02345, DBLP:journals/corr/abs-2510-15047, DBLP:journals/corr/abs-2602-05842}, reflection and action evaluation~\cite{DBLP:journals/corr/abs-2603-08706, DBLP:conf/emnlp/XiaSWYKRYM25, DBLP:journals/corr/abs-2505-20023, DBLP:journals/corr/abs-2501-11425} and domain-specific skill-prior learning~\cite{DBLP:journals/corr/abs-2601-18418, DBLP:journals/corr/abs-2509-23045}.
As a representative, Agent Early Experience~\cite{DBLP:journals/corr/abs-2510-08558} develops two variants for constructing world-modeling and self-reflection data from reward-free interactions respectively.
Unlike these approaches, which mostly turn environment experience into auxiliary supervision for prediction or reflection, \textsc{State2State} turns explored states into end-to-end comprehensive state-reaching RL objectives.


\noindent\textbf{Learning from Environment Exploration.}
A closely related direction leverages environment exploration to generate agent training signals. 
Recent efforts~\cite{DBLP:journals/corr/abs-2511-10395, DBLP:journals/corr/abs-2509-15738, wang2025adapting} collect autonomous interaction experience and transform the resulting trajectories into tasks, demonstrations or world knowledge, often with additional teacher annotation. 
Hindsight-based methods~\cite{DBLP:conf/nips/AndrychowiczCRS17, DBLP:journals/corr/abs-2603-21357, DBLP:conf/icml/PourcelCO25} use achieved outcomes or failed trajectories as alternative goals or supervision. 
These works demonstrate the value of learning from exploration, but they often require model-based annotation to ensure quality, and their generated objectives remain largely guided by human-task distributions. 
In contrast, \textsc{State2State} enables environment learning without external teachers or human-task distributions.

\section{Method}

\subsection{Overview}

We present \textsc{State2State}, an environment-centric mid-training method that turns reachable environment states into verifiable RL objectives for LLM agents. 
The core idea is to derive both the task and reward from the environment itself.
Instead of learning from expert trajectories or externally specified tasks, \textsc{State2State} asks the agent to transform an initial environment state into a target state previously explored. 
Task success is verified by comparing the resulting observation with the target observation, enabling environment-derived reward signals.
This formulation applies to environments where reachable states can be reproduced through initialization or replay, and where observations provide a reliable proxy for verifying task-relevant state equivalence.

As shown in Figure~\ref{fig:framework}, \textsc{State2State} proceeds in three stages. First, an exploration policy interacts with the environment and collects reachable observations, forming a pool of candidate target states. 
Second, we filter, sample, and replay these observations to construct reproducible state-reaching tasks, each specified by an initial environment configuration and a target observation. 
Third, the agent is trained with RL to reach the target observation through environment interaction. 
The resulting policy is then used as initialization for downstream RL on standard human-specified tasks.


\subsection{Environment Exploration}

The first stage of \textsc{State2State} collects reachable states through environment exploration. 
The exploration policy can in principle be an LLM policy, a heuristic policy, or a random policy over current valid actions. 
While stronger LLM explorers may produce more purposeful trajectories, they incur high inference cost and latency. 
Moreover, they may introduce task-oriented priors that bias exploration toward states similar to human tasks and therefore reduce state diversity.
This is misaligned with our goal: we expect exploration to uncover diverse states from the environment, not limited by human task distributions.

Therefore, we deliberately adopt random exploration as a simple, efficient and scalable strategy for state collection. 
At each step, the explorer samples an action from the valid action set derived from the environment and executes it. 
When useful, the sampling distribution can be lightly adjusted with environment-specific priors, such as favoring navigation actions in early steps to enhance exploration breadth and manipulation actions in later steps to prioritize interaction depth. 
Since no LLM inference is required, random exploration can be highly parallelized and scaled primarily with environment interaction, enabling the construction of large candidate pools of target states.


\subsection{\textsc{State2State} Task Construction}

We construct \textsc{State2State} tasks from trajectories collected during environment exploration.
In many interactive environments, the full underlying state is inaccessible or environment-specific, while the agent observes the environment through textual or structured observations.
We therefore use observations as an operational representation of state: a target state is specified by a reachable observation, and task success is verified by matching the current observation against the target.

Given the explored observation pool, we first filter out invalid or uninformative observations, such as error messages and meaningless transitions.
We then sample target observations with a diversity-oriented strategy that favors broad coverage of unique reachable observations while limiting excessive repetition of the same instance.
To ensure that targets define reproducible objectives, we replay the corresponding exploration trajectory from the same initial configuration for three times and retain only observations that can be consistently reached.

Each retained target observation is paired with its source environment configuration to form a \textsc{State2State} task.
At training time, the environment is reset to this initial configuration, and the agent is asked to reach the target observation through interaction.
Formally, let \(o(s)\) denote the observation exposed at environment state \(s\), and let \(\mathcal{O}_{\mathrm{exp}}\) be the set of reachable observations collected during exploration. 
A \textsc{State2State} task is specified by an initial state \(s_0\) and a target observation \(o^\star \in \mathcal{O}_{\mathrm{exp}}\); starting from \(s_0\), the policy succeeds once it reaches a state \(s_t\) with \(\mathrm{match}(o(s_t), o^\star)=1\).

The target observation also naturally serves as a verifier: after each step, we compare the current observation with the target using an environment-specific matching function.
In our textual environments, this is instantiated as normalized exact match, yielding the reward:
\begin{equation}
  r_t =
  \left\{
  \begin{array}{ll}
    1, & \mathrm{match}(o(s_t), o^\star), \\
    0, & \mathrm{otherwise}.
  \end{array}
  \right.
\end{equation}

Unlike standard human tasks, where objectives are given as natural-language instructions and verified by task-specific rewards, \textsc{State2State} derives both objectives and verifiers from reachable environment states.
Solving these tasks requires the agent to infer environment dynamics, plan over long-horizon interactions, and manipulate the environment toward the target observation, encouraging the agent to acquire useful environment skill priors for downstream human tasks.

\subsection{\textsc{State2State} Mid-Training}

We optimize \textsc{State2State} tasks with the GRPO algorithm augmented with dynamic sampling~\cite{DBLP:journals/corr/abs-2503-14476}. 
For each target observation, we sample multiple rollouts from the same initial configuration and optimize them with the GRPO algorithm, which compares successful and failed attempts to provide contrastive rewards. 
Additionally, we apply dynamic sampling by discarding rollout groups with identical rewards and continuing to sample new groups until the required batch size is reached. 
This increases the density of informative groups and adaptively focuses training on targets near the current policy frontier, allowing the model to effectively learn from environment-derived targets that naturally vary in difficulty.

Finally, \textsc{State2State} serves as an environment-centric mid-training stage before downstream RL on human-specified tasks. 
After environment learning, we continue training the policy on human-specified tasks using reinforcement learning, where objectives are natural-language instructions and rewards are given by task-specific verifiers. 
This design separates environment capability acquisition from human-intent alignment: \textsc{State2State} first provides scalable environment-grounded skill priors, while downstream RL adapts these priors to solve human-specified tasks.

\begin{table*}[t]
  \centering
  \setlength{\tabcolsep}{8pt}
  \resizebox{0.9\linewidth}{!}{
  \begin{tabular}{llccccccc}
    \toprule
    \multirow{2}{*}{Model} & \multirow{2}{*}{Method}
    & \multirow{2}{*}{\makecell{Human-Specified\\Task Training}}
    & \multicolumn{3}{c}{ALFWorld} & \multicolumn{3}{c}{ScienceWorld} \\
    \cmidrule(lr){4-6} \cmidrule(lr){7-9}
    & & & ID & OOD & Average & ID & OOD & Average \\
    \midrule
    \multirow{4}{*}{Prompting-based}
    & GPT-5.2 & \xmark & 75.71 & 78.36 & 77.04 & 50.50 & 47.50 & 49.00 \\
    & Claude 4.5 Haiku & \xmark & 79.29 & 81.34 & 80.32 & 48.00 & 48.00 & 48.00 \\
    & Qwen-Plus & \xmark & 80.00 & 85.07 & 82.54 & 44.50 & 45.50 & 45.00 \\
    & DeepSeek V4 Flash & \xmark & 84.29 & 87.31 & 85.80 & 44.50 & 39.00 & 41.75 \\
    \midrule
    \multirow{6}{*}{Qwen3-4B}
    & Base & \xmark & 35.71 & 31.34 & 33.53 & 25.75 & 21.50 & 23.63 \\
    \rowcolor{gray!8}
    & \textsc{State2State} (Ours) & \xmark & 43.57 & 46.27 & 44.92 & 24.50 & 17.50 & 21.00 \\
    \cmidrule(lr){2-9}
    & Agent early experience & \cmark & 67.14 & 76.87 & 72.01 & 41.00 & 41.00 & 41.00 \\
    & Distillation SFT & \cmark & 64.29 & 67.91 & 66.10 & 48.00 & 44.75 & 46.38 \\
    & RL & \cmark & 86.43 & 85.82 & 86.13 & 51.00 & 48.25 & 49.63 \\
    \rowcolor{gray!8}
    & \textsc{State2State} + RL (Ours) & \cmark & \textbf{90.71} & \textbf{93.28} & \textbf{92.00} & \textbf{59.75} & \textbf{51.25} & \textbf{55.50} \\
    \midrule
    \multirow{6}{*}{Qwen3-8B}
    & Base & \xmark & 74.29 & 76.12 & 75.21 & 28.50 & 34.50 & 31.50 \\
    \rowcolor{gray!8}
    & \textsc{State2State} (Ours) & \xmark & 70.00 & 84.33 & 77.17 & 31.75 & 34.50 & 33.13 \\
    \cmidrule(lr){2-9}
    & Agent early experience & \cmark & 70.00 & 72.39 & 71.20 & 43.25 & 44.00 & 43.63 \\
    & Distillation SFT & \cmark & 83.57 & 88.81 & 86.19 & 47.00 & 49.75 & 48.38 \\
    & RL & \cmark & 91.43 & 93.28 & 92.36 & 56.00 & 48.25 & 52.13 \\
    \rowcolor{gray!8}
    & \textsc{State2State} + RL (Ours) & \cmark & \textbf{97.14} & \textbf{97.76} & \textbf{97.45} & \textbf{59.50} & \textbf{52.50} & \textbf{56.00} \\
    \bottomrule
  \end{tabular}
  }
  \caption{Main results on ALFWorld and ScienceWorld, reported as task success rates. Human-Specified Task Training indicates whether the model is further trained on standard human-specified tasks in our experiments. \textsc{State2State} generally improves base models as a standalone environment-learning stage, and the full \textsc{State2State} + RL pipeline achieves the best performance across both benchmarks and model scales.}
  \vspace{-4mm}
  \label{tab:main-results}
\end{table*}

\section{Experiments}

To demonstrate the effectiveness of \textsc{State2State}, we conduct experiments to examine whether \textsc{State2State} can further improve downstream agent RL beyond direct task-specific training. 

\subsection{Benchmarks and Evaluation}

We evaluate \textsc{State2State} on two interactive environments: ScienceWorld~\cite{DBLP:conf/emnlp/WangJCA22} and ALFWorld~\cite{DBLP:conf/iclr/ShridharYCBTH21}.
ScienceWorld focuses on science-oriented environments, where agents must gather observations, perform procedural manipulation, and execute long-horizon action sequences to complete a scientific experiment. 
ALFWorld focuses on household navigation and manipulation in textual embodied environments. 
For both benchmarks, we report task success rates on in-distribution (\textit{ID}) and out-of-distribution (\textit{OOD}) partitions. 
For ScienceWorld, we randomly sample 400 tasks from each partition due to large volume and for ALFWorld, we use the full set. For stable evaluation, we set temperature to 0.1 during model generation. 
We include more details in Appendix~\ref{sec:appendix-envs}.

\subsection{Experiment Setup}

We view \textsc{State2State} as a mid-training stage rather than a compute-matched replacement for downstream RL. 
Therefore, the main comparison tests whether environment-derived mid-training provides useful initialization before task-specific RL.
We conduct experiments with two models of different scales, Qwen3-4B and Qwen3-8B~\cite{DBLP:journals/corr/abs-2505-09388}, under several training settings. 
\textsc{State2State} trains the model solely on environment-derived \textsc{State2State} tasks, which evaluates the transferable interaction capabilities induced by environment learning alone. 
RL-only(denoted as RL) directly optimizes the model on downstream human tasks without mid-training. 
\textsc{State2State} + RL is our full pipeline, where the model first learns environment perception and manipulation, and then continues training on downstream human tasks. 
Comparing this setting with RL-only isolates the effect of \textsc{State2State} as a mid-training stage for downstream agent RL.

For additional reference, we also include two baselines based on expert demonstrations: distillation SFT and agent early experience~\cite{DBLP:journals/corr/abs-2510-08558}, described in Appendix~\ref{sec:appendix-baselines}. 
For SFT, we collect 200 high-quality successful trajectories from Qwen-Plus using rejection sampling. 
Additionally, we report the performance of several competent LLMs including GPT-5.2~\cite{singh2025openai}, Claude 4.5 Haiku~\cite{anthropic2025claudehaiku45}, DeepSeek V4 Flash~\cite{deepseekai2026deepseekv4} and Qwen-Plus~\cite{DBLP:journals/corr/abs-2505-09388}. 
For ScienceWorld, the prompting agents are tested on 200 samples each split to reduce cost.

\subsection{Implementation Details}

We use a random policy to explore both environments. 
For ScienceWorld, which involves larger maps, we adopt a two-stage sampling strategy: early exploration favors movement and observation actions, while later exploration favors scientific and state-changing actions. 
We collect exploration episodes in parallel with hundreds of steps per episode. 
For filtering, we filter invalid actions, ambiguous commands, error messages, and other uninformative observations, and then sample diverse targets by prioritizing unique observations and limiting max repetition number to 5 for each observation. 
Task success is verified by observation matching, and we use textual normalized exact match for both environments. 
We include more details about our constructed dataset and several representative examples in Appendix~\ref{sec:appendix-datasets}.

We train \textsc{State2State} with GRPO augmented with dynamic sampling for 80 steps and continue downstream training with the best-performing checkpoint on the state-reaching validation set. 
We present more details in Appendix~\ref{sec:appendix-prompts}.

\subsection{Main Results}

\noindent\textbf{Standalone environment learning.}
Table~\ref{tab:main-results} reports the main results on ALFWorld and ScienceWorld. 
As a standalone environment-learning stage, \textsc{State2State} improves performance in most settings, with consistent gains on ALFWorld and additional gains for Qwen3-8B on ScienceWorld. 
This shows that environment-centric \textsc{State2State} can induce useful environment cognition and manipulation capabilities even before training on human-specified tasks. 
The only exception is Qwen3-4B on ScienceWorld with slight degradation, possibly due to the sparse and challenging exact state-reaching objectives for smaller models. 
Nevertheless, the same mid-training stage substantially improves downstream RL, indicating that the learned environment priors are better realized after human-task alignment.

\noindent\textbf{Stronger initialization for downstream RL.}
The full \textsc{State2State} + RL pipeline achieves the best performance across both environments and model scales. 
Its gains over direct RL appear not only on ALFWorld, where RL is already strong, but also on the more challenging ScienceWorld benchmark. 
Together with consistent improvements on both ID and OOD splits, these results suggest that \textsc{State2State} provides useful environment-level capabilities rather than merely fitting a specific task distribution. 
The resulting models also surpass the strong prompting LLMs in these environments.

\begin{figure}[t]
  \centering
  \includegraphics[width=\linewidth]{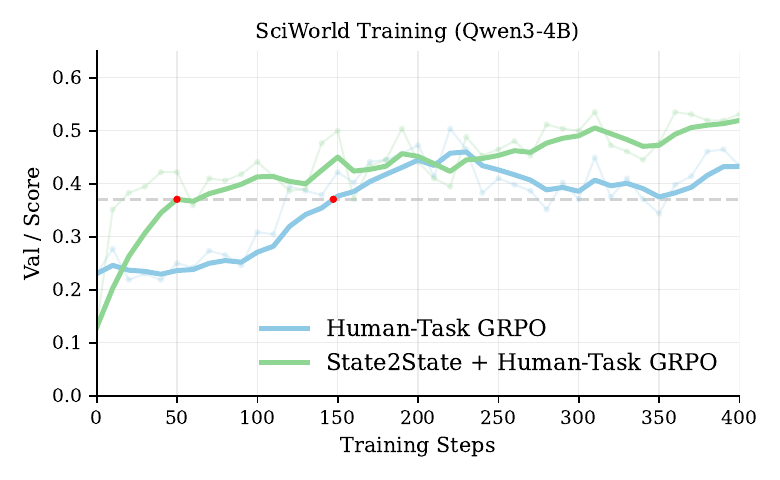}
  \vspace{-3mm}
  \caption{Training curves for Qwen3-4B on ScienceWorld validation set. Ours achieves stronger final performance and higher learning efficiency.}
  \label{fig:ScienceWorld-training-curve}
  \vspace{-4mm}
\end{figure}

\noindent\textbf{Improved learning efficiency.}
Figure~\ref{fig:ScienceWorld-training-curve} further shows that \textsc{State2State} improves not only final performance but also downstream RL efficiency. 
On ScienceWorld with Qwen3-4B, \textsc{State2State} + Human-Task GRPO reaches a relatively high score by around step 50, comparable to the score reached by Human-Task GRPO at around step 150.
This indicates that environment-level mid-training provides useful skill priors that can accelerate subsequent learning on human tasks.

\noindent\textbf{Latent benefits released by human-task RL.}
The full pipeline yields more consistent gains over direct RL than \textsc{State2State}-only training over the base model. Especially for Qwen3-8B, adding \textsc{State2State} before RL brings larger improvements than applying \textsc{State2State} alone on both ALFWorld ($+5.09$ vs. $+1.96$) and ScienceWorld ($+3.87$ vs. $+1.63$), suggesting that \textsc{State2State} learns reusable environment-level capabilities whose benefits are only partially reflected by state-reaching performance alone. 
Downstream RL can further align these capabilities with user-facing task goals, thereby converting latent environment manipulation skills into stronger task-solving performance.

\section{Analysis}

We conduct the following analyses to better understand the mechanisms of \textsc{State2State}, with experiments on ScienceWorld with Qwen3-4B model unless otherwise specified.

\subsection{Ordering of SFT and \textsc{State2State}}

We analyze how \textsc{State2State} interacts with commonly used SFT method using the same dataset for the distillation SFT baseline. 
When applying \textsc{State2State} on the stronger SFT model, we observe more consistent and stable performance gains and extend the \textsc{State2State} training to 150 steps.
As shown in Table~\ref{tab:sft-rl-analysis}, applying \textsc{State2State} after SFT substantially improves over SFT, showing that \textsc{State2State} objectives derived from environment explorations still provide complementary benefits beyond SFT.

\begin{table}[t]
  \centering
  \small
  \setlength{\tabcolsep}{6pt}
  \resizebox{0.8\linewidth}{!}{%
  \begin{tabular}{lccc}
    \toprule
    Method & ID & OOD & Avg. \\
    \midrule
    Base & 25.75 & 21.50 & 23.63 \\
    SFT & 48.00 & 44.75 & 46.38 \\
    \textsc{S2S} + SFT & 47.25 & 48.75 & 48.00 \\
    SFT + \textsc{S2S} & 55.75 & 52.50 & 54.13 \\
    \midrule
    SFT + RL & 65.00 & 51.75 & 58.38 \\
    \textsc{S2S} + SFT + RL & 60.75 & 60.25 & 60.50 \\
    SFT + \textsc{S2S} + RL & \textbf{70.75} & \textbf{61.00} & \textbf{65.88} \\
    \bottomrule
  \end{tabular}
  }
  \caption{Ordering of SFT and \textsc{State2State}, with \textsc{State2State} denoted as \textsc{S2S}. SFT + \textsc{S2S} + RL performs best, showing that \textsc{S2S} is most effective as a mid-training stage placed after SFT and before RL.}
  \vspace{-3mm}
  \label{tab:sft-rl-analysis}
\end{table}

The benefit becomes clearer when followed by downstream RL.
SFT + \textsc{State2State} + RL achieves the best overall and OOD performance, indicating that \textsc{State2State} remains effective on top of SFT, benefiting from a stronger initial policy.
This suggests that expert demonstrations and environment learning are complementary: while SFT imitates human-task trajectories, \textsc{State2State} trains on state-reaching objectives sampled from a broader reachable environment space.
The training order is also important: applying \textsc{State2State} before SFT brings limited gains, likely because subsequent imitation learning can overwrite part of the exploratory environment priors. 
Therefore, \textsc{State2State} is best used after SFT and before final task-specific RL, where it can improve the SFT policy with environment-grounded skill priors.

\subsection{Integration with Different RL Backbones}

We examine whether \textsc{State2State} remains effective with a stronger downstream RL backbone algorithm. 
We use GiGPO~\cite{DBLP:journals/corr/abs-2505-10978}, an RL algorithm with fine-grained rewards designed for agentic RL, and compare each backbone with and without \textsc{State2State} mid-training.

\begin{table}[t]
  \centering
  \small
  \setlength{\tabcolsep}{6pt}
  \resizebox{0.9\linewidth}{!}{%
  \begin{tabular}{lccc}
    \toprule
    Method & ID & OOD & Avg. \\
    \midrule
    Base & 25.75 & 21.50 & 23.63 \\
    GRPO & 51.00 & 48.25 & 49.63 \\
    \textsc{State2State} + GRPO & 59.75 & 51.25 & 55.50 \\
    GiGPO & 60.25 & 49.25 & 54.75 \\
    \textsc{State2State} + GiGPO & \textbf{61.75} & \textbf{53.75} & \textbf{57.75} \\
    \bottomrule
  \end{tabular}
  }
  \caption{Analysis of different RL backbone algorithms on ScienceWorld. \textsc{State2State} improves both GRPO and GiGPO, showing that its benefit is complementary to the downstream optimizer.}
  \label{tab:gigpo-analysis}
  \vspace{-6mm}
\end{table}

As shown in Table~\ref{tab:gigpo-analysis}, \textsc{State2State} + GRPO reaches 55.50 average, outperforming GiGPO alone, showing that environment learning with a standard RL backbone can already match or exceed a stronger RL algorithm without \textsc{State2State}. 
When combined with GiGPO, \textsc{State2State} further improves the performance on both ID and OOD partitions. 
These results indicate that \textsc{State2State} is complementary to the choice of RL algorithm and can provide consistent benefits even with a stronger RL optimizer.

\subsection{Effect of Exploration Strategy}

The environment exploration stage can instantiate different exploration policies for collecting target states. 
We conduct experiments with two choices: LLM exploration and random exploration. 
For LLM exploration, we use Qwen-Plus as the strong LLM explorer. 
As Table~\ref{tab:exploration-analysis} shows, both variants improve over direct RL, showing that \textsc{State2State} can benefit from different sources of states.

\begin{table}[t]
  \centering
  \small
  \setlength{\tabcolsep}{6pt}
  \resizebox{\linewidth}{!}{%
  \begin{tabular}{llccc}
    \toprule
    Method & Exploration & ID & OOD & Avg. \\
    \midrule
    Base & -- & 25.75 & 21.50 & 23.63 \\
    RL & -- & 51.00 & 48.25 & 49.63 \\
    \textsc{State2State} + RL & LLM & 54.75 & 50.50 & 52.63 \\
    \textsc{State2State} + RL & Random & \textbf{59.75} & \textbf{51.25} & \textbf{55.50} \\
    \bottomrule
  \end{tabular}
  }
  \caption{Analysis of exploration strategies on ScienceWorld. Random exploration yields the strongest \textsc{State2State} + RL result, showing that low-cost state discovery can provide effective training targets.}
  \label{tab:exploration-analysis}
\end{table}

Random exploration performs better compared with LLM explorers, and we attribute this to the greater depth and diversity of random exploration in the current environments. 
Based on our observations, since LLM explorers tend to carry priors toward manually designed tasks, they may visit a narrower region of the reachable state space, while also requiring higher inference cost and latency. 
In contrast, random exploration provides a cheaper and broader way to collect diverse target states. 

\subsection{Cross-Environment Generalization}

To examine the generalization of \textsc{State2State}, we evaluate cross-environment transfer from ScienceWorld to ALFWorld. 
Specifically, we first conduct \textsc{State2State} mid-training on ScienceWorld and then perform downstream RL on ALFWorld. 
We also compare against a variant that uses ScienceWorld human-task RL before ALFWorld RL, with the same amount of mid-training steps.

\begin{table}[t]
  \centering
  \small
  \setlength{\tabcolsep}{6pt}
  \resizebox{0.9\linewidth}{!}{%
  \begin{tabular}{llrrr}
    \toprule
    Mid Training &  ID & OOD & Avg. \\
    \midrule
    None &  86.43 & 85.82 & 86.13 \\
    ScienceWorld RL &  86.43 & 87.31 & 86.87 \\
    ScienceWorld \textsc{State2State} &  \textbf{88.57} & \textbf{90.30} & \textbf{89.44} \\
    \bottomrule
  \end{tabular}
  }
  \caption{Cross-environment transfer results from ScienceWorld to ALFWorld, showing that \textsc{State2State} as mid-training transfers better than human tasks.
  Before training on ALFWorld standard tasks, we conduct mid-training on ScienceWorld with \textsc{State2State} tasks and human-specified tasks separately.}
  \vspace{-4mm}
  \label{tab:generalization-analysis}
\end{table}

As shown in Table~\ref{tab:generalization-analysis}, ScienceWorld \textsc{State2State} improves performance over the baseline without mid-training, indicating that \textsc{State2State} can provide positive transfer across environments with related dynamics. 
In contrast, RL on ScienceWorld human tasks shows limited capability transfer.
Since this baseline uses the same steps of mid-training, the stronger transfer of our method suggests that the \textsc{State2State} objective, rather than additional cross-environment RL alone, is critical for learning generalizable environment capabilities.

Overall, these results suggest that \textsc{State2State} supports both in-domain adaptation and potentially promising cross-environment transfer.

\begin{table}[t]
  \centering
  \small
  \resizebox{0.75\linewidth}{!}{%
  \begin{tabular}{lcc}
    \toprule
    Method & Training Steps & Score \\
    \midrule
    MAI-UI-8B & 0  & 0.275 \\
    \textsc{State2State} & 80  & \textbf{0.308} \\
    \bottomrule
  \end{tabular}
  }
  \caption{Results on the GUI-only subset of MobileWorld. \textsc{State2State} improves MAI-UI-8B in a mobile GUI environment without downstream RL on human-specified tasks.}
  \vspace{-4mm}
  \label{tab:mobileworld}
\end{table}

\subsection{Extension to Mobile GUI Environments}

To examine whether \textsc{State2State} can extend beyond text-based environments, we conduct an additional experiment on MobileWorld~\cite{DBLP:journals/corr/abs-2512-19432}, a more complex mobile GUI benchmark with long-horizon and cross-application tasks.
We evaluate on its GUI-only subset, containing 117 complex tasks.
MobileWorld provides a suitable testbed for evaluating standalone \textsc{State2State}: its benchmark tasks are limited without available official training data, and scaling human-specified task training with reliable verifiers is costly.
We therefore evaluate \textsc{State2State} without downstream RL on human-specified tasks, directly testing whether environment-derived objectives can improve interaction capabilities in a more complex GUI environment.
More details are in Appendix~\ref{sec:appendix-mobileworld}.

As shown in Table~\ref{tab:mobileworld}, \textsc{State2State} improves MAI-UI-8B from 0.275 to 0.308 after 80 training steps.
This result suggests that \textsc{State2State} can provide useful environment-learning signals in more complex GUI environments, where collecting human-specified training tasks is difficult.

\section{Conclusion}

We present \textsc{State2State}, an environment-derived mid-training method that converts explored environment states into verifiable state-reaching objectives. 
By deriving both objectives and rewards from the environment, \textsc{State2State} enables agents to acquire environment skill priors without expert supervision or human-specified tasks. 
Across several environments, \textsc{State2State} standalone generally improves agent performance, strengthens downstream human-task RL, and shows promising cross-environment generalization.

\section*{Limitations}

Our experiments provide an initial study of environment learning for LLM agents, prioritizing controlled online RL comparisons over exhaustive scaling. 
We evaluate \textsc{State2State} on Qwen3-4B and Qwen3-8B, which are strong enough to exhibit meaningful interactive learning but do not cover frontier-scale models. 
Larger models may interact with \textsc{State2State} differently due to stronger priors, longer-horizon planning, or better exploration. 
While the state-reaching objectives and rule-based verification should remain applicable, systematic scaling analysis is needed to understand how gains evolve with model size and training compute.

Our evaluation also focuses on representative environments with reproducible states and verifiable success conditions. 
ALFWorld and ScienceWorld test household and scientific procedural interaction, and our results suggest benefits for both in-domain adaptation and cross-environment transfer. 
However, environment learning may benefit from richer reachable state spaces. 
Extending \textsc{State2State} to more visually grounded, web-based, software-engineering, and real-device control environments would better assess the generality of the learned manipulation capabilities. 
We view this as a natural extension, since the core method only requires reproducible states and a state-matching verifier.

\bibliography{custom}

@inproceedings{DBLP:conf/acl/ZengLLWLD024,
  author       = {Aohan Zeng and
                  Mingdao Liu and
                  Rui Lu and
                  Bowen Wang and
                  Xiao Liu and
                  Yuxiao Dong and
                  Jie Tang},
  editor       = {Lun{-}Wei Ku and
                  Andre Martins and
                  Vivek Srikumar},
  title        = {AgentTuning: Enabling Generalized Agent Abilities for LLMs},
  booktitle    = {Findings of the Association for Computational Linguistics, {ACL} 2024,
                  Bangkok, Thailand and virtual meeting, August 11-16, 2024},
  series       = {Findings of {ACL}},
  pages        = {3053--3077},
  publisher    = {Association for Computational Linguistics},
  year         = {2024},
  url          = {https://doi.org/10.18653/v1/2024.findings-acl.181},
  doi          = {10.18653/V1/2024.FINDINGS-ACL.181},
  bibsource    = {dblp computer science bibliography, https://dblp.org}
}

@article{DBLP:journals/corr/abs-2310-05915,
  author       = {Baian Chen and
                  Chang Shu and
                  Ehsan Shareghi and
                  Nigel Collier and
                  Karthik Narasimhan and
                  Shunyu Yao},
  title        = {FireAct: Toward Language Agent Fine-tuning},
  journal      = {CoRR},
  volume       = {abs/2310.05915},
  year         = {2023},
  url          = {https://doi.org/10.48550/arXiv.2310.05915},
  doi          = {10.48550/ARXIV.2310.05915},
  eprinttype   = {arXiv},
  eprint       = {2310.05915},
  bibsource    = {dblp computer science bibliography, https://dblp.org}
}

@article{DBLP:journals/corr/abs-2510-24701,
  author       = {Baixuan Li and
                  Bo Zhang and
                  Dingchu Zhang and
                  Fei Huang and
                  Guangyu Li and
                  Guoxin Chen and
                  Huifeng Yin and
                  Jialong Wu and
                  Jingren Zhou and
                  Kuan Li and
                  Liangcai Su and
                  Litu Ou and
                  Liwen Zhang and
                  Pengjun Xie and
                  Rui Ye and
                  Wenbiao Yin and
                  Xinmiao Yu and
                  Xinyu Wang and
                  Xixi Wu and
                  Xuanzhong Chen and
                  Yida Zhao and
                  Zhen Zhang and
                  Zhengwei Tao and
                  Zhongwang Zhang and
                  Zile Qiao and
                  Chenxi Wang and
                  Donglei Yu and
                  Gang Fu and
                  Haiyang Shen and
                  Jiayin Yang and
                  Jun Lin and
                  Junkai Zhang and
                  Kui Zeng and
                  Li Yang and
                  Hailong Yin and
                  Maojia Song and
                  Ming Yan and
                  Peng Xia and
                  Qian Xiao and
                  Rui Min and
                  Ruixue Ding and
                  Runnan Fang and
                  Shaowei Chen and
                  Shen Huang and
                  Shihang Wang and
                  Shihao Cai and
                  Weizhou Shen and
                  Xiaobin Wang and
                  Xin Guan and
                  Xinyu Geng and
                  Yingcheng Shi and
                  Yuning Wu and
                  Zhuo Chen and
                  Zijian Li and
                  Yong Jiang},
  title        = {Tongyi DeepResearch Technical Report},
  journal      = {CoRR},
  volume       = {abs/2510.24701},
  year         = {2025},
  url          = {https://doi.org/10.48550/arXiv.2510.24701},
  doi          = {10.48550/ARXIV.2510.24701},
  eprinttype   = {arXiv},
  eprint       = {2510.24701},
  bibsource    = {dblp computer science bibliography, https://dblp.org}
}

@article{DBLP:journals/corr/abs-2507-20534,
  author       = {Kimi Team},
  title        = {Kimi {K2:} Open Agentic Intelligence},
  journal      = {CoRR},
  volume       = {abs/2507.20534},
  year         = {2025},
  url          = {https://doi.org/10.48550/arXiv.2507.20534},
  doi          = {10.48550/ARXIV.2507.20534},
  eprinttype   = {arXiv},
  eprint       = {2507.20534},
  bibsource    = {dblp computer science bibliography, https://dblp.org}
}

@article{DBLP:journals/corr/abs-2509-13310,
  author       = {Liangcai Su and
                  Zhen Zhang and
                  Guangyu Li and
                  Zhuo Chen and
                  Chenxi Wang and
                  Maojia Song and
                  Xinyu Wang and
                  Kuan Li and
                  Jialong Wu and
                  Xuanzhong Chen and
                  Zile Qiao and
                  Zhongwang Zhang and
                  Huifeng Yin and
                  Shihao Cai and
                  Runnan Fang and
                  Zhengwei Tao and
                  Wenbiao Yin and
                  Chenxiong Qian and
                  Yong Jiang and
                  Pengjun Xie and
                  Fei Huang and
                  Jingren Zhou},
  title        = {Scaling Agents via Continual Pre-training},
  journal      = {CoRR},
  volume       = {abs/2509.13310},
  year         = {2025},
  url          = {https://doi.org/10.48550/arXiv.2509.13310},
  doi          = {10.48550/ARXIV.2509.13310},
  eprinttype   = {arXiv},
  eprint       = {2509.13310},
  bibsource    = {dblp computer science bibliography, https://dblp.org}
}

@article{DBLP:journals/corr/abs-2508-09123,
  author       = {Xinyuan Wang and
                  Bowen Wang and
                  Dunjie Lu and
                  Junlin Yang and
                  Tianbao Xie and
                  Junli Wang and
                  Jiaqi Deng and
                  Xiaole Guo and
                  Yiheng Xu and
                  Chen Henry Wu and
                  Zhennan Shen and
                  Zhuokai Li and
                  Ryan Li and
                  Xiaochuan Li and
                  Junda Chen and
                  Boyuan Zheng and
                  Peihang Li and
                  Fangyu Lei and
                  Ruisheng Cao and
                  Yeqiao Fu and
                  Dongchan Shin and
                  Martin Shin and
                  Jiarui Hu and
                  Yuyan Wang and
                  Jixuan Chen and
                  Yuxiao Ye and
                  Danyang Zhang and
                  Dikang Du and
                  Hao Hu and
                  Huarong Chen and
                  Zaida Zhou and
                  Haotian Yao and
                  Ziwei Chen and
                  Qizheng Gu and
                  Yipu Wang and
                  Heng Wang and
                  Diyi Yang and
                  Victor Zhong and
                  Flood Sung and
                  Y. Charles and
                  Zhilin Yang and
                  Tao Yu},
  title        = {OpenCUA: Open Foundations for Computer-Use Agents},
  journal      = {CoRR},
  volume       = {abs/2508.09123},
  year         = {2025},
  url          = {https://doi.org/10.48550/arXiv.2508.09123},
  doi          = {10.48550/ARXIV.2508.09123},
  eprinttype   = {arXiv},
  eprint       = {2508.09123},
  bibsource    = {dblp computer science bibliography, https://dblp.org}
}

@inproceedings{DBLP:conf/iclr/QiLILSSYYY00D25,
  author       = {Zehan Qi and
                  Xiao Liu and
                  Iat Long Iong and
                  Hanyu Lai and
                  Xueqiao Sun and
                  Jiadai Sun and
                  Xinyue Yang and
                  Yu Yang and
                  Shuntian Yao and
                  Wei Xu and
                  Jie Tang and
                  Yuxiao Dong},
  title        = {WebRL: Training {LLM} Web Agents via Self-Evolving Online Curriculum
                  Reinforcement Learning},
  booktitle    = {The Thirteenth International Conference on Learning Representations,
                  {ICLR} 2025, Singapore, April 24-28, 2025},
  publisher    = {OpenReview.net},
  year         = {2025},
  url          = {https://openreview.net/forum?id=oVKEAFjEqv},
  bibsource    = {dblp computer science bibliography, https://dblp.org}
}

@article{DBLP:journals/corr/abs-2402-03300,
  author       = {Zhihong Shao and
                  Peiyi Wang and
                  Qihao Zhu and
                  Runxin Xu and
                  Junxiao Song and
                  Mingchuan Zhang and
                  Y. K. Li and
                  Y. Wu and
                  Daya Guo},
  title        = {DeepSeekMath: Pushing the Limits of Mathematical Reasoning in Open
                  Language Models},
  journal      = {CoRR},
  volume       = {abs/2402.03300},
  year         = {2024},
  url          = {https://doi.org/10.48550/arXiv.2402.03300},
  doi          = {10.48550/ARXIV.2402.03300},
  eprinttype   = {arXiv},
  eprint       = {2402.03300},
  bibsource    = {dblp computer science bibliography, https://dblp.org}
}

@article{DBLP:journals/corr/abs-2504-20073,
  author       = {Zihan Wang and
                  Kangrui Wang and
                  Qineng Wang and
                  Pingyue Zhang and
                  Linjie Li and
                  Zhengyuan Yang and
                  Xing Jin and
                  Kefan Yu and
                  Minh Nhat Nguyen and
                  Licheng Liu and
                  Eli Gottlieb and
                  Yiping Lu and
                  Kyunghyun Cho and
                  Jiajun Wu and
                  Li Fei{-}Fei and
                  Lijuan Wang and
                  Yejin Choi and
                  Manling Li},
  title        = {{RAGEN:} Understanding Self-Evolution in {LLM} Agents via Multi-Turn
                  Reinforcement Learning},
  journal      = {CoRR},
  volume       = {abs/2504.20073},
  year         = {2025},
  url          = {https://doi.org/10.48550/arXiv.2504.20073},
  doi          = {10.48550/ARXIV.2504.20073},
  eprinttype   = {arXiv},
  eprint       = {2504.20073},
  bibsource    = {dblp computer science bibliography, https://dblp.org}
}

@article{DBLP:journals/corr/abs-2505-10978,
  author       = {Lang Feng and
                  Zhenghai Xue and
                  Tingcong Liu and
                  Bo An},
  title        = {Group-in-Group Policy Optimization for {LLM} Agent Training},
  journal      = {CoRR},
  volume       = {abs/2505.10978},
  year         = {2025},
  url          = {https://doi.org/10.48550/arXiv.2505.10978},
  doi          = {10.48550/ARXIV.2505.10978},
  eprinttype   = {arXiv},
  eprint       = {2505.10978},
  bibsource    = {dblp computer science bibliography, https://dblp.org}
}

@article{DBLP:journals/corr/abs-2507-22844,
  author       = {Zijing Zhang and
                  Ziyang Chen and
                  Mingxiao Li and
                  Zhaopeng Tu and
                  Xiaolong Li},
  title        = {{RLVMR:} Reinforcement Learning with Verifiable Meta-Reasoning Rewards
                  for Robust Long-Horizon Agents},
  journal      = {CoRR},
  volume       = {abs/2507.22844},
  year         = {2025},
  url          = {https://doi.org/10.48550/arXiv.2507.22844},
  doi          = {10.48550/ARXIV.2507.22844},
  eprinttype   = {arXiv},
  eprint       = {2507.22844},
  bibsource    = {dblp computer science bibliography, https://dblp.org}
}

@article{DBLP:journals/corr/abs-2503-09516,
  author       = {Bowen Jin and
                  Hansi Zeng and
                  Zhenrui Yue and
                  Dong Wang and
                  Hamed Zamani and
                  Jiawei Han},
  title        = {Search-R1: Training LLMs to Reason and Leverage Search Engines with
                  Reinforcement Learning},
  journal      = {CoRR},
  volume       = {abs/2503.09516},
  year         = {2025},
  url          = {https://doi.org/10.48550/arXiv.2503.09516},
  doi          = {10.48550/ARXIV.2503.09516},
  eprinttype   = {arXiv},
  eprint       = {2503.09516},
  bibsource    = {dblp computer science bibliography, https://dblp.org}
}

@article{DBLP:journals/corr/abs-2510-23081,
  author       = {Chengying Tu and
                  Xuemiao Zhang and
                  Rongxiang Weng and
                  Rumei Li and
                  Chen Zhang and
                  Yang Bai and
                  Hongfei Yan and
                  Jingang Wang and
                  Xunliang Cai},
  title        = {A Survey on {LLM} Mid-training},
  journal      = {CoRR},
  volume       = {abs/2510.23081},
  year         = {2025},
  url          = {https://doi.org/10.48550/arXiv.2510.23081},
  doi          = {10.48550/ARXIV.2510.23081},
  eprinttype   = {arXiv},
  eprint       = {2510.23081},
  bibsource    = {dblp computer science bibliography, https://dblp.org}
}

@article{DBLP:journals/corr/abs-2510-08558,
  author       = {Kai Zhang and
                  Xiangchao Chen and
                  Bo Liu and
                  Tianci Xue and
                  Zeyi Liao and
                  Zhihan Liu and
                  Xiyao Wang and
                  Yuting Ning and
                  Zhaorun Chen and
                  Xiaohan Fu and
                  Jian Xie and
                  Yuxuan Sun and
                  Boyu Gou and
                  Qi Qi and
                  Zihang Meng and
                  Jianwei Yang and
                  Ning Zhang and
                  Xian Li and
                  Ashish Shah and
                  Dat Huynh and
                  Hengduo Li and
                  Zi Yang and
                  Sara Cao and
                  Lawrence Jang and
                  Shuyan Zhou and
                  Jiacheng Zhu and
                  Huan Sun and
                  Jason Weston and
                  Yu Su and
                  Yifan Wu},
  title        = {Agent Learning via Early Experience},
  journal      = {CoRR},
  volume       = {abs/2510.08558},
  year         = {2025},
  url          = {https://doi.org/10.48550/arXiv.2510.08558},
  doi          = {10.48550/ARXIV.2510.08558},
  eprinttype   = {arXiv},
  eprint       = {2510.08558},
  bibsource    = {dblp computer science bibliography, https://dblp.org}
}

@article{DBLP:journals/corr/abs-2603-08706,
  author       = {Weize Liu and
                  Minghui Liu and
                  Sy{-}Tuyen Ho and
                  Souradip Chakraborty and
                  Xiyao Wang and
                  Furong Huang},
  title        = {Agentic Critical Training},
  journal      = {CoRR},
  volume       = {abs/2603.08706},
  year         = {2026},
  url          = {https://doi.org/10.48550/arXiv.2603.08706},
  doi          = {10.48550/ARXIV.2603.08706},
  eprinttype   = {arXiv},
  eprint       = {2603.08706},
  bibsource    = {dblp computer science bibliography, https://dblp.org}
}

@inproceedings{DBLP:conf/emnlp/XiaSWYKRYM25,
  author       = {Yu Xia and
                  Yiran Shen and
                  Junda Wu and
                  Tong Yu and
                  Sungchul Kim and
                  Ryan A. Rossi and
                  Lina Yao and
                  Julian J. McAuley},
  editor       = {Christos Christodoulopoulos and
                  Tanmoy Chakraborty and
                  Carolyn Rose and
                  Violet Peng},
  title        = {{SAND:} Boosting {LLM} Agents with Self-Taught Action Deliberation},
  booktitle    = {Proceedings of the 2025 Conference on Empirical Methods in Natural
                  Language Processing, {EMNLP} 2025, Suzhou, China, November 4-9, 2025},
  pages        = {3062--3077},
  publisher    = {Association for Computational Linguistics},
  year         = {2025},
  url          = {https://doi.org/10.18653/v1/2025.emnlp-main.152},
  doi          = {10.18653/V1/2025.EMNLP-MAIN.152},
  bibsource    = {dblp computer science bibliography, https://dblp.org}
}

@article{DBLP:journals/corr/abs-2604-02345,
  author       = {Mengzhou Wu and
                  Yuzhe Guo and
                  Yuan Cao and
                  Haochuan Lu and
                  Songhe Zhu and
                  Pingzhe Qu and
                  Xin Chen and
                  Kang Qin and
                  Zhongpu Wang and
                  Xiaode Zhang and
                  Xinyi Wang and
                  Wei Dai and
                  Gang Cao and
                  Yuetang Deng and
                  Zhi Gong and
                  Dezhi Ran and
                  Linyi Li and
                  Wei Yang and
                  Tao Xie},
  title        = {UI-Oceanus: Scaling {GUI} Agents with Synthetic Environmental Dynamics},
  journal      = {CoRR},
  volume       = {abs/2604.02345},
  year         = {2026},
  url          = {https://doi.org/10.48550/arXiv.2604.02345},
  doi          = {10.48550/ARXIV.2604.02345},
  eprinttype   = {arXiv},
  eprint       = {2604.02345},
  bibsource    = {dblp computer science bibliography, https://dblp.org}
}

@article{DBLP:journals/corr/abs-2601-18418,
  author       = {Ji Zeng and
                  Dayuan Fu and
                  Tiantian Mi and
                  Yumin Zhuang and
                  Yaxing Huang and
                  Xuefeng Li and
                  Lyumanshan Ye and
                  Muhang Xie and
                  Qishuo Hua and
                  Zhen Huang and
                  Mohan Jiang and
                  Hanning Wang and
                  Jifan Lin and
                  Yang Xiao and
                  Jie Sun and
                  Yunze Wu and
                  Pengfei Liu},
  title        = {daVinci-Dev: Agent-native Mid-training for Software Engineering},
  journal      = {CoRR},
  volume       = {abs/2601.18418},
  year         = {2026},
  url          = {https://doi.org/10.48550/arXiv.2601.18418},
  doi          = {10.48550/ARXIV.2601.18418},
  eprinttype   = {arXiv},
  eprint       = {2601.18418},
  bibsource    = {dblp computer science bibliography, https://dblp.org}
}

@article{DBLP:journals/corr/abs-2509-23045,
  author       = {Zonghan Yang and
                  Shengjie Wang and
                  Kelin Fu and
                  Wenyang He and
                  Weimin Xiong and
                  Yibo Liu and
                  Yibo Miao and
                  Bofei Gao and
                  Yejie Wang and
                  Yingwei Ma and
                  Yanhao Li and
                  Yue Liu and
                  Zhenxing Hu and
                  Kaitai Zhang and
                  Shuyi Wang and
                  Huarong Chen and
                  Flood Sung and
                  Yang Liu and
                  Yang Gao and
                  Zhilin Yang and
                  Tianyu Liu},
  title        = {Kimi-Dev: Agentless Training as Skill Prior for SWE-Agents},
  journal      = {CoRR},
  volume       = {abs/2509.23045},
  year         = {2025},
  url          = {https://doi.org/10.48550/arXiv.2509.23045},
  doi          = {10.48550/ARXIV.2509.23045},
  eprinttype   = {arXiv},
  eprint       = {2509.23045},
  bibsource    = {dblp computer science bibliography, https://dblp.org}
}

@article{DBLP:journals/corr/abs-2510-15047,
  author       = {Shiqi Chen and
                  Tongyao Zhu and
                  Zian Wang and
                  Jinghan Zhang and
                  Kangrui Wang and
                  Siyang Gao and
                  Teng Xiao and
                  Yee Whye Teh and
                  Junxian He and
                  Manling Li},
  title        = {Internalizing World Models via Self-Play Finetuning for Agentic {RL}},
  journal      = {CoRR},
  volume       = {abs/2510.15047},
  year         = {2025},
  url          = {https://doi.org/10.48550/arXiv.2510.15047},
  doi          = {10.48550/ARXIV.2510.15047},
  eprinttype   = {arXiv},
  eprint       = {2510.15047},
  bibsource    = {dblp computer science bibliography, https://dblp.org}
}

@article{DBLP:journals/corr/abs-2602-05842,
  author       = {Xiao Yu and
                  Baolin Peng and
                  Ruize Xu and
                  Yelong Shen and
                  Pengcheng He and
                  Suman Nath and
                  Nikhil Singh and
                  Jiangfeng Gao and
                  Zhou Yu},
  title        = {Reinforcement World Model Learning for LLM-based Agents},
  journal      = {CoRR},
  volume       = {abs/2602.05842},
  year         = {2026},
  url          = {https://doi.org/10.48550/arXiv.2602.05842},
  doi          = {10.48550/ARXIV.2602.05842},
  eprinttype   = {arXiv},
  eprint       = {2602.05842},
  bibsource    = {dblp computer science bibliography, https://dblp.org}
}

@article{DBLP:journals/corr/abs-2505-20023,
  author       = {Yihan Chen and
                  Benfeng Xu and
                  Xiaorui Wang and
                  Yongdong Zhang and
                  Zhendong Mao},
  title        = {Training LLM-Based Agents with Synthetic Self-Reflected Trajectories
                  and Partial Masking},
  journal      = {CoRR},
  volume       = {abs/2505.20023},
  year         = {2025},
  url          = {https://doi.org/10.48550/arXiv.2505.20023},
  doi          = {10.48550/ARXIV.2505.20023},
  eprinttype   = {arXiv},
  eprint       = {2505.20023},
  bibsource    = {dblp computer science bibliography, https://dblp.org}
}

@article{DBLP:journals/corr/abs-2501-11425,
  author       = {Siyu Yuan and
                  Zehui Chen and
                  Zhiheng Xi and
                  Junjie Ye and
                  Zhengyin Du and
                  Jiecao Chen},
  title        = {Agent-R: Training Language Model Agents to Reflect via Iterative Self-Training},
  journal      = {CoRR},
  volume       = {abs/2501.11425},
  year         = {2025},
  url          = {https://doi.org/10.48550/arXiv.2501.11425},
  doi          = {10.48550/ARXIV.2501.11425},
  eprinttype   = {arXiv},
  eprint       = {2501.11425},
  bibsource    = {dblp computer science bibliography, https://dblp.org}
}

@article{DBLP:journals/corr/abs-2511-10395,
  author       = {Yunpeng Zhai and
                  Shuchang Tao and
                  Cheng Chen and
                  Anni Zou and
                  Ziqian Chen and
                  Qingxu Fu and
                  Shinji Mai and
                  Li Yu and
                  Jiaji Deng and
                  Zouying Cao and
                  Zhaoyang Liu and
                  Bolin Ding and
                  Jingren Zhou},
  title        = {AgentEvolver: Towards Efficient Self-Evolving Agent System},
  journal      = {CoRR},
  volume       = {abs/2511.10395},
  year         = {2025},
  url          = {https://doi.org/10.48550/arXiv.2511.10395},
  doi          = {10.48550/ARXIV.2511.10395},
  eprinttype   = {arXiv},
  eprint       = {2511.10395},
  bibsource    = {dblp computer science bibliography, https://dblp.org}
}

@article{DBLP:journals/corr/abs-2509-15738,
  author       = {Musen Lin and
                  Minghao Liu and
                  Taoran Lu and
                  Lichen Yuan and
                  Yiwei Liu and
                  Haonan Xu and
                  Yu Miao and
                  Yuhao Chao and
                  Zhaojian Li},
  title        = {GUI-ReWalk: Massive Data Generation for {GUI} Agent via Stochastic
                  Exploration and Intent-Aware Reasoning},
  journal      = {CoRR},
  volume       = {abs/2509.15738},
  year         = {2025},
  url          = {https://doi.org/10.48550/arXiv.2509.15738},
  doi          = {10.48550/ARXIV.2509.15738},
  eprinttype   = {arXiv},
  eprint       = {2509.15738},
  bibsource    = {dblp computer science bibliography, https://dblp.org}
}

@article{wang2025adapting,
  title={Adapting Web Agents with Synthetic Supervision},
  author={Wang, Zhaoyang and Liang, Yiming and Zhang, Xuchao and Wu, Qianhui and Han, Siwei and Bastos, Anson and Wang, Rujia and Bansal, Chetan and Peng, Baolin and Gao, Jianfeng and others},
  journal={arXiv preprint arXiv:2511.06101},
  year={2025}
}

@inproceedings{DBLP:conf/nips/AndrychowiczCRS17,
  author       = {Marcin Andrychowicz and
                  Dwight Crow and
                  Alex Ray and
                  Jonas Schneider and
                  Rachel Fong and
                  Peter Welinder and
                  Bob McGrew and
                  Josh Tobin and
                  Pieter Abbeel and
                  Wojciech Zaremba},
  editor       = {Isabelle Guyon and
                  Ulrike von Luxburg and
                  Samy Bengio and
                  Hanna M. Wallach and
                  Rob Fergus and
                  S. V. N. Vishwanathan and
                  Roman Garnett},
  title        = {Hindsight Experience Replay},
  booktitle    = {Advances in Neural Information Processing Systems 30: Annual Conference
                  on Neural Information Processing Systems 2017, December 4-9, 2017,
                  Long Beach, CA, {USA}},
  pages        = {5048--5058},
  year         = {2017},
  url          = {https://proceedings.neurips.cc/paper/2017/hash/453fadbd8a1a3af50a9df4df899537b5-Abstract.html},
  bibsource    = {dblp computer science bibliography, https://dblp.org}
}

@inproceedings{DBLP:conf/icml/PourcelCO25,
  author       = {Julien Pourcel and
                  C{\'{e}}dric Colas and
                  Pierre{-}Yves Oudeyer},
  editor       = {Aarti Singh and
                  Maryam Fazel and
                  Daniel Hsu and
                  Simon Lacoste{-}Julien and
                  Felix Berkenkamp and
                  Tegan Maharaj and
                  Kiri Wagstaff and
                  Jerry Zhu},
  title        = {Self-Improving Language Models for Evolutionary Program Synthesis:
                  {A} Case Study on {ARC-AGI}},
  booktitle    = {Forty-second International Conference on Machine Learning, {ICML}
                  2025, Vancouver, BC, Canada, July 13-19, 2025},
  series       = {Proceedings of Machine Learning Research},
  publisher    = {{PMLR} / OpenReview.net},
  year         = {2025},
  url          = {https://proceedings.mlr.press/v267/pourcel25a.html},
  bibsource    = {dblp computer science bibliography, https://dblp.org}
}

@article{DBLP:journals/corr/abs-2603-21357,
  author       = {Liang Ding},
  title        = {AgentHER: Hindsight Experience Replay for {LLM} Agent Trajectory Relabeling},
  journal      = {CoRR},
  volume       = {abs/2603.21357},
  year         = {2026},
  url          = {https://doi.org/10.48550/arXiv.2603.21357},
  doi          = {10.48550/ARXIV.2603.21357},
  eprinttype   = {arXiv},
  eprint       = {2603.21357},
  bibsource    = {dblp computer science bibliography, https://dblp.org}
}

@inproceedings{DBLP:conf/iclr/ShridharYCBTH21,
  author       = {Mohit Shridhar and
                  Xingdi Yuan and
                  Marc{-}Alexandre C{\^{o}}t{\'{e}} and
                  Yonatan Bisk and
                  Adam Trischler and
                  Matthew J. Hausknecht},
  title        = {ALFWorld: Aligning Text and Embodied Environments for Interactive
                  Learning},
  booktitle    = {9th International Conference on Learning Representations, {ICLR} 2021,
                  Virtual Event, Austria, May 3-7, 2021},
  publisher    = {OpenReview.net},
  year         = {2021},
  url          = {https://openreview.net/forum?id=0IOX0YcCdTn},
  bibsource    = {dblp computer science bibliography, https://dblp.org}
}

@inproceedings{DBLP:conf/emnlp/WangJCA22,
  author       = {Ruoyao Wang and
                  Peter A. Jansen and
                  Marc{-}Alexandre C{\^{o}}t{\'{e}} and
                  Prithviraj Ammanabrolu},
  editor       = {Yoav Goldberg and
                  Zornitsa Kozareva and
                  Yue Zhang},
  title        = {ScienceWorld: Is your Agent Smarter than a 5th Grader?},
  booktitle    = {Proceedings of the 2022 Conference on Empirical Methods in Natural
                  Language Processing, {EMNLP} 2022, Abu Dhabi, United Arab Emirates,
                  December 7-11, 2022},
  pages        = {11279--11298},
  publisher    = {Association for Computational Linguistics},
  year         = {2022},
  url          = {https://doi.org/10.18653/v1/2022.emnlp-main.775},
  doi          = {10.18653/V1/2022.EMNLP-MAIN.775},
  bibsource    = {dblp computer science bibliography, https://dblp.org}
}

@article{DBLP:journals/corr/abs-2506-10055,
  author       = {Dingfeng Shi and
                  Jingyi Cao and
                  Qianben Chen and
                  Weichen Sun and
                  Weizhen Li and
                  Hongxuan Lu and
                  Fangchen Dong and
                  Tianrui Qin and
                  King Zhu and
                  Minghao Liu and
                  Jian Yang and
                  Ge Zhang and
                  Jiaheng Liu and
                  Changwang Zhang and
                  Jun Wang and
                  Yuchen Eleanor Jiang and
                  Wangchunshu Zhou},
  title        = {TaskCraft: Automated Generation of Agentic Tasks},
  journal      = {CoRR},
  volume       = {abs/2506.10055},
  year         = {2025},
  url          = {https://doi.org/10.48550/arXiv.2506.10055},
  doi          = {10.48550/ARXIV.2506.10055},
  eprinttype   = {arXiv},
  eprint       = {2506.10055},
  bibsource    = {dblp computer science bibliography, https://dblp.org}
}

@inproceedings{DBLP:conf/kdd/Hu0XSLLLR25,
  author       = {Mengkang Hu and
                  Pu Zhao and
                  Can Xu and
                  Qingfeng Sun and
                  Jian{-}Guang Lou and
                  Qingwei Lin and
                  Ping Luo and
                  Saravan Rajmohan},
  editor       = {Yizhou Sun and
                  Flavio Chierichetti and
                  Hady W. Lauw and
                  Claudia Perlich and
                  Wee Hyong Tok and
                  Andrew Tomkins},
  title        = {AgentGen: Enhancing Planning Abilities for Large Language Model based
                  Agent via Environment and Task Generation},
  booktitle    = {Proceedings of the 31st {ACM} {SIGKDD} Conference on Knowledge Discovery
                  and Data Mining, V.1, {KDD} 2025, Toronto, ON, Canada, August 3-7,
                  2025},
  pages        = {496--507},
  publisher    = {{ACM}},
  year         = {2025},
  url          = {https://doi.org/10.1145/3690624.3709321},
  doi          = {10.1145/3690624.3709321},
  bibsource    = {dblp computer science bibliography, https://dblp.org}
}

@article{DBLP:journals/corr/abs-2512-19432,
  author       = {Quyu Kong and
                  Xu Zhang and
                  Zhenyu Yang and
                  Nolan Gao and
                  Chen Liu and
                  Panrong Tong and
                  Chenglin Cai and
                  Hanzhang Zhou and
                  Jianan Zhang and
                  Liangyu Chen and
                  Zhidan Liu and
                  Steven Hoi and
                  Yue Wang},
  title        = {MobileWorld: Benchmarking Autonomous Mobile Agents in Agent-User Interactive
                  and MCP-Augmented Environments},
  journal      = {CoRR},
  volume       = {abs/2512.19432},
  year         = {2025},
  url          = {https://doi.org/10.48550/arXiv.2512.19432},
  doi          = {10.48550/ARXIV.2512.19432},
  eprinttype   = {arXiv},
  eprint       = {2512.19432},
  bibsource    = {dblp computer science bibliography, https://dblp.org}
}

@article{DBLP:journals/corr/abs-2503-14476,
  author       = {Qiying Yu and
                  Zheng Zhang and
                  Ruofei Zhu and
                  Yufeng Yuan and
                  Xiaochen Zuo and
                  Yu Yue and
                  Tiantian Fan and
                  Gaohong Liu and
                  Lingjun Liu and
                  Xin Liu and
                  Haibin Lin and
                  Zhiqi Lin and
                  Bole Ma and
                  Guangming Sheng and
                  Yuxuan Tong and
                  Chi Zhang and
                  Mofan Zhang and
                  Wang Zhang and
                  Hang Zhu and
                  Jinhua Zhu and
                  Jiaze Chen and
                  Jiangjie Chen and
                  Chengyi Wang and
                  Hongli Yu and
                  Weinan Dai and
                  Yuxuan Song and
                  Xiangpeng Wei and
                  Hao Zhou and
                  Jingjing Liu and
                  Wei{-}Ying Ma and
                  Ya{-}Qin Zhang and
                  Lin Yan and
                  Mu Qiao and
                  Yonghui Wu and
                  Mingxuan Wang},
  title        = {{DAPO:} An Open-Source {LLM} Reinforcement Learning System at Scale},
  journal      = {CoRR},
  volume       = {abs/2503.14476},
  year         = {2025},
  url          = {https://doi.org/10.48550/arXiv.2503.14476},
  doi          = {10.48550/ARXIV.2503.14476},
  eprinttype   = {arXiv},
  eprint       = {2503.14476},
  bibsource    = {dblp computer science bibliography, https://dblp.org}
}

@article{DBLP:journals/corr/abs-2505-09388,
  author       = {Qwen Team},
  title        = {Qwen3 Technical Report},
  journal      = {CoRR},
  volume       = {abs/2505.09388},
  year         = {2025},
  url          = {https://doi.org/10.48550/arXiv.2505.09388},
  doi          = {10.48550/ARXIV.2505.09388},
  eprinttype   = {arXiv},
  eprint       = {2505.09388},
  bibsource    = {dblp computer science bibliography, https://dblp.org}
}

@article{singh2025openai,
  title={Openai gpt-5 system card},
  author={Singh, Aaditya and Fry, Adam and Perelman, Adam and Tart, Adam and Ganesh, Adi and El-Kishky, Ahmed and McLaughlin, Aidan and Low, Aiden and Ostrow, AJ and Ananthram, Akhila and others},
  journal={arXiv preprint arXiv:2601.03267},
  year={2025}
}

@misc{deepseekai2026deepseekv4,
  title        = {DeepSeek-V4: Towards Highly Efficient Million-Token Context Intelligence},
  author       = {DeepSeek-AI},
  year         = {2026},
  url          = {https://huggingface.co/deepseek-ai/DeepSeek-V4-Pro/blob/main/DeepSeek_V4.pdf}
}

@misc{anthropic2025claudehaiku45,
  title        = {System Card: Claude Haiku 4.5},
  author       = {Anthropic},
  year         = {2025},
  url          = {https://www-cdn.anthropic.com/7aad69bf12627d42234e01ee7c36305dc2f6a970.pdf}
}

@article{sheng2024hybridflow,
  title   = {HybridFlow: A Flexible and Efficient RLHF Framework},
  author  = {Guangming Sheng and Chi Zhang and Zilingfeng Ye and Xibin Wu and Wang Zhang and Ru Zhang and Yanghua Peng and Haibin Lin and Chuan Wu},
  year    = {2024},
  journal = {arXiv preprint arXiv: 2409.19256}
}

@inproceedings{kwon2023efficient,
  title={Efficient Memory Management for Large Language Model Serving with PagedAttention},
  author={Woosuk Kwon and Zhuohan Li and Siyuan Zhuang and Ying Sheng and Lianmin Zheng and Cody Hao Yu and Joseph E. Gonzalez and Hao Zhang and Ion Stoica},
  booktitle={Proceedings of the ACM SIGOPS 29th Symposium on Operating Systems Principles},
  year={2023}
}

\newpage

\appendix

\section{Environments}
\label{sec:appendix-envs}

\subsection{ALFWorld}
ALFWorld~\cite{DBLP:conf/iclr/ShridharYCBTH21} is a text-based embodied household environment built on ALFRED and TextWorld.
In the standard human tasks, each episode starts from a household scene and a task instruction, such as finding, cleaning, heating, cooling, examining, or placing objects.
At each step, the agent receives a textual observation describing the consequence of its previous actions.
The environment also exposes a set of admissible commands, which are used for environment exploration by the random policy.
The agent must output a command inside an \texttt{<action>} span, and the command is executed by the simulator.

We use the ALFWorld text environment for standard human task training and evaluation.
Episodes terminate according to the environment success signal, and the reward returned from the environment is \(1.0\) for successful completion and \(0.0\) otherwise.

The \textsc{State2State} training reuses the same environment as the standard human tasks but replaces the original natural-language task instruction with a target observation as the state-reaching objective.
Each task specifies a source game file for environment initialization and a target observation string.
After each action, the current textual observation is compared with the target observation after normalization.
The episode succeeds and terminates when the normalized observation exactly matches the normalized target.
Thus, \textsc{State2State} provides a binary, rule-based state-matching reward without using the original ALFWorld task success verifier.

Training and evaluation are run with parallel environment workers.
For grouped RL, repeated workers in the same group load the same task instance, enabling multiple rollouts from the same initial configuration.

\subsection{ScienceWorld}

ScienceWorld~\cite{DBLP:conf/emnlp/WangJCA22} is a text-based interactive environment for elementary-level science tasks.
Each episode presents a scientific goal that may require navigation, object inspection, object manipulation, container operations, device activation, and experiment execution.
The agent receives a textual observation at each step and must issue one valid action.
The action space is built from action templates, possible objects, and valid action-object combinations exposed by the simulator. In the prompt, we present the generic action templates.

In the standard ScienceWorld environment, each task is specified by a task name and variation.
The simulator maintains a task score in the range from 0 to 100.
In our reward design, we reward only perfect completion of the task: score 100 from the environment will receive a \(1.0\) for successful completion and \(0.0\) otherwise.
In our main experiments, ScienceWorld evaluation reports success rates on sampled in-distribution and out-of-distribution task partitions.

The \textsc{State2State} training uses the same ScienceWorld environments but defines the objective as reaching a target observation string, without access to the original environment task instruction.
Each \textsc{State2State} task JSON provides the original task name, task variation, and \texttt{task\_target\_state}.
After each action, the current observation is compared against the target observation using exact string equality under normalization.
If the strings match, the episode terminates with reward \(1\); otherwise the reward remains \(0\).
This setting turns explored ScienceWorld observations into verifiable state-reaching objectives.

\section{\textsc{State2State} Datasets}
\label{sec:appendix-datasets}

This section describes the construction of the \textsc{State2State} datasets used for environment-derived mid-training.
Both datasets are generated from environment exploration rather than expert demonstrations or human-written task instructions.
The resulting examples specify a reproducible environment configuration and a target observation, and success is verified by matching the current observation to the target observation.

\subsection{ScienceWorld Dataset}

\paragraph{Exploration.}
The ScienceWorld \textsc{State2State} dataset is built from random exploration trajectories generated with 2560 exploration episodes.
The exploration seed is 42, and each episode allows up to 300 environment steps.

The exploration policy samples from environment actions after several validity and compatibility filters.
Invalid actions, non-actionable objects, and semantically incompatible action-object pairs are removed before sampling.
Because ScienceWorld has comparatively large maps, exploration uses an explore-then-operate two-phase strategy: a spatial-exploration phase ending at 20\% of the episode, and during the rest 80\% steps we encourage the scientific and state-changing actions.

\paragraph{Processing and filtering.}
The dataset construction process loads all exploration trajectories and collects observations from each trajectory's set of unique observations.
For each observation, the pipeline records provenance information including source trajectory, source step, task name, task variation, seed, and exploration strategy.
It then removes invalid or uninformative observations using a keyword blacklist and a short-text heuristic for error-like messages.

\paragraph{Sampling.}
ScienceWorld target selection follows a diversity-first procedure.
The first phase ensures that each unique valid observation appears at least once.
If more tasks are needed, later phases add additional instances subject to a maximum observation frequency of 5.
The target number is 8,000 tasks.
Each exported task contains a task identifier, the original ScienceWorld task name and variation, a \texttt{task\_target\_state} field, and metadata describing the source trajectory and step.

\paragraph{Statistics.}
The final ScienceWorld dataset contains 8,000 tasks, split into 7,872 training tasks and 128 validation tasks.
All 8,000 target observations are unique, giving a diversity ratio of 1.0.
The largest task-name groups are \texttt{test-conductivity} (2,900), \texttt{inclined-plane-friction-named-surfaces} (1,630), \texttt{find-living-thing} (674), \texttt{find-non-living-thing} (574), and \texttt{measure-melting-point-known-substance} (540).
Target observations have median length 72 characters, 90th percentile length 500, 95th percentile length 811, and maximum length 1,276.
The source-step distribution has median 116, 90th percentile 185, 99th percentile 199, and maximum 200.

\subsection{ALFWorld Dataset}

\paragraph{Exploration.}
The ALFWorld \textsc{State2State} dataset is built from random exploration over the ALFWorld training task list.
The production run uses seed 42, 256 parallel environments and 20 episode batches, yielding 5,120 trajectories.
Each trajectory allows up to 500 steps.
The random explorer samples from the environment-provided admissible commands.
Since the distribution of each action verb can be very uneven, we adopt a verb-weighted sampling mechanism: commands are grouped by inferred verb, assigned per-verb weights, and normalized by the number of admissible commands for that verb.

\paragraph{Processing and filtering.}
Candidate target states are collected from all trajectory steps whose index is at least 10; no upper step bound is applied.
Observations are filtered with substring bans for welcome banners and error-like feedback, including messages such as unknown actions, invalid actions, and impossible commands.
For each candidate, the selection script infers a verb from observation text using a regular expression matching phrases of the form \texttt{You <verb>}.

\paragraph{Deduplication and sampling.}
To encourage the state-changing actions, we use different deduplication scopes for different verb categories in ALFWorld.
For state-changing verbs like \texttt{clean}, \texttt{cool} and \texttt{heat}, duplicate observations are removed only within the same trajectory, preserving similar outcomes reached in different initial games.
For all other verbs, duplicate observations are removed globally across trajectories.
The final dataset is sampled without replacement from the candidate pool using verb-weighted sampling, with preference towards state-changing actions like heat or cool.
Each exported task contains a task identifier, a source game file for environment initialization, and a \texttt{task\_target\_state} field.

\paragraph{Statistics.}
The final ALFWorld dataset contains 4,256 tasks, split into 4,000 training tasks and 256 validation tasks.
The final verb distribution is: \texttt{arrive} (733), \texttt{clean} (435), \texttt{cool} (423), \texttt{move} (423), \texttt{turn} (396), \texttt{pick} (381), \texttt{open} (361), \texttt{sliced} (320), \texttt{see} (280), \texttt{are} (206), \texttt{heat} (204), \texttt{other} (51), and \texttt{close} (43).
Target observations are shorter than those in ScienceWorld: the median length is 44 characters, the 90th percentile is 104, the 95th percentile is 135, and the maximum is 360.

\begin{table*}[t]
  \centering
  \small
  \setlength{\tabcolsep}{6pt}
  \resizebox{0.75\linewidth}{!}{
  \begin{tabular}{lcccccc}
    \toprule
    Dataset & Tasks & Train & Val & Traj. &  Median Len. & Max Len. \\
    \midrule
    ScienceWorld & 8,000 & 7,872 & 128 & 2,560  & 72 & 1,276 \\
    ALFWorld & 4,256 & 4,000 & 256 & 5,120  & 44 & 360 \\
    \bottomrule
  \end{tabular}
  }
  \caption{Summary statistics for the \textsc{State2State} datasets.}
  \label{tab:appendix-dataset-stats}
\end{table*}

\subsection{Representative Cases}

The following examples show representative \textsc{State2State} targets sampled from the case files in \texttt{resources/appendix\_2}.
They are chosen to cover different task families, observation types, and action effects.

\begin{tcolorbox}[breakable, colback=bg_rl, colframe=border_gray, title=\textbf{ScienceWorld Case 1: \texttt{boil}; room inventory}, fonttitle=\bfseries\small, boxrule=0.5mm]
\begin{lstlisting}
This room is called the greenhouse. 
In it, you see:
the agent
a substance called air
a bee hive. The bee hive door is open. In the bee hive is: nothing.
a sink, which is turned on. In the sink is: the ground, a hole (containing a substance called water).
soil
You also see:
A door to the hallway (that is open)
A door to the outside (that is open)
\end{lstlisting}
\end{tcolorbox}

\begin{tcolorbox}[breakable, colback=bg_rl, colframe=border_gray, title=\textbf{ScienceWorld Case 2: \texttt{power-component}; inventory}, fonttitle=\bfseries\small, boxrule=0.5mm]
\begin{lstlisting}
In your inventory, you see:
a blue light bulb, which is off. 
its anode is connected to: a terminal 1 on agent. 
its cathode is connected to: a terminal 2 on agent.
a green light bulb, which is off. 
its anode is connected to: a terminal 1 on door. 
its cathode is connected to: a terminal 1 on drain.
a substance called wood
\end{lstlisting}
\end{tcolorbox}

\begin{tcolorbox}[breakable, colback=bg_rl, colframe=border_gray, title=\textbf{ScienceWorld Case 3: \texttt{find-animal}; action feedback}, fonttitle=\bfseries\small, boxrule=0.5mm]
\begin{lstlisting}
You dunk the flower pot 9 in the flower pot 5, moving the liquids (water, water, water, water, water) to the flower pot 9.
\end{lstlisting}
\end{tcolorbox}

\begin{tcolorbox}[breakable, colback=bg_rl, colframe=border_gray, title=\textbf{ALFWorld Case 1: \texttt{arrive}; navigation/listing}, fonttitle=\bfseries\small, boxrule=0.5mm]
\begin{lstlisting}
You arrive at countertop 1. 
On the countertop 1, 
you see a bowl 2, a bread 3, 
a bread 2, a bread 1, 
a fork 1, a houseplant 1,
a kettle 2, a knife 2, a knife 1.
\end{lstlisting}
\end{tcolorbox}

\begin{tcolorbox}[breakable, colback=bg_rl, colframe=border_gray, title=\textbf{ALFWorld Case 2: \texttt{open}; container state}, fonttitle=\bfseries\small, boxrule=0.5mm]
\begin{lstlisting}
You open the fridge 1. 
The fridge 1 is open. 
In it, you see a apple 2, a bread 1, 
a egg 2, a egg 1, a mug 1, 
a plate 2, a potato 1, 
a tomato 2, and a tomato 1.
\end{lstlisting}
\end{tcolorbox}

\begin{tcolorbox}[breakable, colback=bg_rl, colframe=border_gray, title=\textbf{ALFWorld Case 3: \texttt{heat}; state-changing action}, fonttitle=\bfseries\small, boxrule=0.5mm]
\begin{lstlisting}
You heat the potato 1 
using the microwave 1.
\end{lstlisting}
\end{tcolorbox}

\section{RL Algorithms}
\label{sec:appendix-rl-algorithms}

This section briefly summarizes the RL algorithms used in our experiments.
All of them are group-based policy optimization methods: for each task or target state, the policy samples multiple rollouts under the same initial condition, and learning is driven by relative performance within the group.

\subsection{GRPO}

Group Relative Policy Optimization (GRPO)~\cite{DBLP:journals/corr/abs-2402-03300} is the main RL backbone used in our \textsc{State2State} and downstream human-task training experiments.
Compared with PPO, GRPO removes the learned value function and estimates the baseline from a group of sampled responses or trajectories for the same prompt.
In our agent setting, for a task \(x\), we sample \(G\) rollouts \(\{\tau_i\}_{i=1}^{G}\) from the policy and obtain environment returns \(\{R_i\}_{i=1}^{G}\).
The group-relative advantage is computed by normalizing returns inside the group:
\[
  \hat{A}_i = \frac{R_i - \mathrm{mean}(\{R_j\}_{j=1}^{G})}
  {\mathrm{std}(\{R_j\}_{j=1}^{G})}.
\]
The policy is then updated with a clipped importance-ratio objective, with the same rollout-level advantage assigned to the generated tokens/actions in that rollout.
This critic-free design is well suited to sparse verifiable rewards, because successful and failed attempts on the same task provide a direct contrastive signal.

\subsection{DAPO Dynamic Sampling}

DAPO~\cite{DBLP:journals/corr/abs-2503-14476} introduces several modifications to GRPO for large-scale RL, and in this work we only use the dynamic sampling mechanism.
The motivation is that a rollout group with identical outcomes gives no useful group-relative gradient: if all \(G\) rollouts fail or all \(G\) rollouts succeed, the group provides no meaningful relative comparison.
Dynamic sampling therefore over-samples rollout groups and keeps only informative groups before an update.
For binary verifiable rewards, this corresponds to retaining groups satisfying
\[
  0 < \sum_{i=1}^{G}\mathbf{1}[R_i > 0] < G.
\]
In our implementation, \textsc{State2State} training repeatedly samples groups until enough non-identical outcome groups are collected, up to a maximum generation-batch budget.
This increases the density of effective comparisons for sparse state-reaching rewards.

\subsection{GiGPO}

Group-in-Group Policy Optimization (GiGPO)~\cite{DBLP:journals/corr/abs-2505-10978} is designed for multi-turn LLM agent training, where a single terminal reward may be too coarse for assigning credit to intermediate actions.
GiGPO preserves the episode-level grouping used by GRPO, but adds a second, step-level grouping mechanism.
After collecting multiple trajectories for the same task, it identifies repeated environment states across trajectories as anchor states and groups the actions taken from the same anchor state.
This allows the algorithm to compare locally different actions that were chosen under the same state condition, without requiring extra per-state rollouts.

GiGPO combines an episode-level advantage \(A^E(\tau_i)\) with a step-level advantage \(A^S(a_t^{(i)})\):
\[
  A(a_t^{(i)}) = A^E(\tau_i) + \omega A^S(a_t^{(i)}),
\]
where \(\omega\) controls the strength of step-level credit assignment.
The resulting advantage is used in a clipped policy-gradient objective similar to GRPO.
In this paper, GiGPO is used as an alternative downstream RL backbone in the analysis section to test whether \textsc{State2State} remains beneficial when the optimizer already provides more fine-grained agent credit assignment.

\section{Baselines}
\label{sec:appendix-baselines}

We include two trajectory-based baselines that use expert-style supervision rather than online RL from \textsc{State2State} objectives.

\subsection{Distillation SFT}

The Distillation SFT baseline collects demonstration trajectories by running a strong closed-source LLM Qwen-Plus in ALFWorld and ScienceWorld with the same environment interface used during training.
At each step, the model receives the full prompt constructed by the environment manager, including the task description, current observation, recent interaction history, and admissible actions, and returns a response containing thoughts and an \texttt{<action>} span.
We record the prompt and response for each active step, keep successful trajectories, and convert the resulting step-level pairs into supervised fine-tuning data format. 

\subsection{Agent Early Experience}

Agent Early Experience (AEE)~\cite{DBLP:journals/corr/abs-2510-08558} augments expert trajectories with step-level self-reflection supervision. In our experiment, we adopt the self-reflection variant.
For each expert step, AEE reconstructs the environment state by replaying the expert prefix, samples \(K\) alternative admissible actions excluding the expert action, and executes the expert and alternative actions in parallel environment copies.
An LLM then observes the original situation, the expert action and next state, and an alternative action and next state, and writes a self-reflection explaining why the expert action is preferable.
The final SFT example trains the model to output the reflection followed by the expert \texttt{<action>} tag.
In our experiments, AEE serves as a stronger trajectory-based baseline that provides additional reasoning supervision beyond direct action imitation.

\section{Training Setups and Prompts}
\label{sec:appendix-prompts}

This section presents the key parameters and prompts used for training.
In all settings, the agent is expected to reason and then emit an executable action inside \texttt{<action>} and \texttt{</action>} tags.
Curly-braced fields are filled by the environment manager at rollout time.
Our standard training prompts mainly follow the previous work~\cite{DBLP:journals/corr/abs-2505-10978} with slight modifications. The \textsc{State2State} training prompts are modified from the standard training prompts.

\subsection{Key RL Training Hyperparameters}

All RL training runs, including \textsc{State2State} mid-training and downstream standard-task training, share most optimization settings across model sizes and phases. We adopt verl~\cite{sheng2024hybridflow} open-source RL framework and vllm~\cite{kwon2023efficient} inference engine for high-throughput RL training.
We optimize the actor with learning rate \(1e^{-6}\), enable KL loss with coefficient \(0.01\), and do not add KL penalties directly into the environment reward.
We use GRPO-style RL algorithm in both training phases. We set the train batch size to 16 and the GRPO group size is set to 8. We deploy the environment via ray framework and enable the vllm rollout server to interact with the environments in parallel.
For validation rollouts, we use sampling with temperature \(0.1\) for stability, and for training rollouts, we set the temperature to 1.0 for balanced exploration and exploitation.

For ALFWorld training, we use a maximum prompt length of 4096 and response length of 1024.
For ScienceWorld training, we use longer contexts, with maximum prompt length between 12000 and response length 1024.
\textsc{State2State} runs use dynamic group filtering to continually collect informative rollout groups. We set the maximum number of generation batches to 4 for ALFWorld and ScienceWorld Qwen3-8B \textsc{State2State} training, and to 6 for ScienceWorld Qwen3-4B training.
For the \textsc{State2State} training, we train the model for 80 steps maximum. And for standard task training, we train the model for 400 steps in ScienceWorld and 300 steps ALFWorld, determined by the convergence steps of the baseline training (directly RL on the environments without \textsc{State2State}).
As for the max\_steps in a single episode, we set max\_steps to be 40 for ALFWorld and 30 for ScienceWorld.

\subsection{Standard ALFWorld Training Prompt}

\begin{tcolorbox}[breakable, colback=bg_rl, colframe=border_gray, title=\textbf{Standard ALFWorld Training Prompt}, fonttitle=\bfseries\small, boxrule=0.5mm]
\begin{lstlisting}
You are an expert agent operating in the ALFRED Embodied Environment. Your task is to: {task_description}
Prior to this step, you have already taken {step_count} step(s). Below are the most recent {history_length} observations and the corresponding actions you took: {action_history}
You are now at step {current_step} and your current observation is: {current_observation}
Your admissible actions of the current situation are: [{admissible_actions}].

Now it's your turn to take an action.
You should first reason step-by-step about the current situation.
Once you've finished your reasoning, you should choose an admissible action for current step and present it within <action> </action> tags.
\end{lstlisting}
\end{tcolorbox}

\subsection{\textsc{State2State} Training Prompt}

\begin{tcolorbox}[breakable, colback=bg_rl, colframe=border_gray, title=\textbf{\textsc{State2State} Training Prompt}, fonttitle=\bfseries\small, boxrule=0.5mm]
\begin{lstlisting}
You are an expert agent operating in the ALFRED Embodied Environment.
Your goal is to reach the following TARGET STATE (the environment observation text must match it):
{target_state}
You need to perform actions to make the environment observation the same as the given target state. After your each action, the environment observation will be updated. Only if the environment observation is exactly the same as the target state, the task is considered successful.

Prior to this step, you have already taken {step_count} step(s). Below are the most recent {history_length} observations and the corresponding actions you took: {action_history}
You are now at step {current_step} and your current observation is: {current_observation}
Your admissible actions of the current situation are: [{admissible_actions}].

The target state is: {target_state}

Now it's your turn to take an action.
You should first reason step-by-step about the current situation. This reasoning process MUST be enclosed within <think> </think> tags.
Once you've finished your reasoning, you should choose an admissible action for current step and present it within <action> </action> tags.
\end{lstlisting}
\end{tcolorbox}

\subsection{Standard ScienceWorld Training Prompt}

\begin{tcolorbox}[breakable, colback=bg_rl, colframe=border_gray, title=\textbf{Standard ScienceWorld Training Prompt}, fonttitle=\bfseries\small, boxrule=0.5mm]
\begin{lstlisting}
You are an expert agent operating in the ScienceWorld environment, which is a text-based virtual environment centered around accomplishing tasks from the elementary science curriculum.

Your current task is: {task_description}

Prior to this step, you have already taken {step_count} step(s). Below are the most recent {history_length} observations and the corresponding actions you took: {action_history}
You are now at step {current_step} and your current observation is: {current_observation}

Here are the actions you may take:
[
{"action": "open OBJ", "description": "open a container"},
{"action": "close OBJ", "description": "close a container"},
{"action": "activate OBJ", "description": "activate a device"},
{"action": "deactivate OBJ", "description": "deactivate a device"},
{"action": "connect OBJ to OBJ", "description": "connect electrical components"},
{"action": "disconnect OBJ to OBJ", "description": "disconnect electrical components"},
{"action": "use OBJ [on OBJ]", "description": "use a device/item"},
{"action": "look around", "description": "describe the current room"},
{"action": "look at OBJ", "description": "describe an object in detail"},
{"action": "look in OBJ", "description": "describe a container's contents"},
{"action": "read OBJ", "description": "read a note or book"},
{"action": "move OBJ to OBJ", "description": "move an object to a container"},
{"action": "pick up OBJ", "description": "move an object to the inventory"},
{"action": "put down OBJ", "description": "drop an inventory item"},
{"action": "pour OBJ into OBJ", "description": "pour a liquid into a container"},
{"action": "dunk OBJ into OBJ", "description": "dunk a container into a liquid"},
{"action": "mix OBJ", "description": "chemically mix a container"},
{"action": "go to LOC", "description": "move to a new location"},
{"action": "eat OBJ", "description": "eat a food"},
{"action": "flush OBJ", "description": "flush a toilet"},
{"action": "focus on OBJ", "description": "signal intent on a task object"},
{"action": "wait", "description": "take no action for 10 iterations"},
{"action": "wait1", "description": "take no action for 1 iteration"},
{"action": "task", "description": "describe current task"},
{"action": "inventory", "description": "list your inventory"}
]

Current available actions:
{available_actions}

Tips:
1. You can use look around to see the surroundings of the current location.
2. All containers in the environment have already been opened, you can directly get items from the containers.
3. You can explore the environment and find the items you need to complete the experiment.
**4. Your thinking process should be concise, do not waste time on irrelevant details.**

Your current task is: {task_description}

You are now at step {current_step} and your current observation is: {current_observation}

Now it's your turn to take an action. You should first reason step-by-step about the current situation.
Once you've finished your reasoning, you should choose an appropriate action for the current step and present it within <action> </action> tags.
\end{lstlisting}
\end{tcolorbox}

\subsection{\textsc{State2State} Training Prompt}

\begin{tcolorbox}[breakable, colback=bg_rl, colframe=border_gray, title=\textbf{\textsc{State2State} Training Prompt}, fonttitle=\bfseries\small, boxrule=0.5mm]
\begin{lstlisting}
You are an expert agent operating in the ScienceWorld environment, which is a text-based virtual environment centered around accomplishing tasks from the elementary science curriculum.
Your task is to reach a target state(the environment observation). If the target state is reached, the task is considered successful. If the task is not completed within a certain number of steps, the task is considered failed.

The target state is: {target_state}

You need to perform actions to make the environment observation the same as the given target state. After your each action, the environment observation will be updated. Only if the environment observation is exactly the same as the target state, the task is considered successful.

Prior to this step, you have already taken {step_count} step(s). Below are the most recent {history_length} observations and the corresponding actions you took: {action_history}
You are now at step {current_step} and your current observation is: {current_observation}

Here are the actions you may take:
[
{"action": "open OBJ", "description": "open a container"},
{"action": "close OBJ", "description": "close a container"},
{"action": "activate OBJ", "description": "activate a device"},
{"action": "deactivate OBJ", "description": "deactivate a device"},
{"action": "connect OBJ to OBJ", "description": "connect electrical components"},
{"action": "disconnect OBJ to OBJ", "description": "disconnect electrical components"},
{"action": "use OBJ [on OBJ]", "description": "use a device/item"},
{"action": "look around", "description": "describe the current room"},
{"action": "look at OBJ", "description": "describe an object in detail"},
{"action": "look in OBJ", "description": "describe a container's contents"},
{"action": "read OBJ", "description": "read a note or book"},
{"action": "move OBJ to OBJ", "description": "move an object to a container"},
{"action": "pick up OBJ", "description": "move an object to the inventory"},
{"action": "put down OBJ", "description": "drop an inventory item"},
{"action": "pour OBJ into OBJ", "description": "pour a liquid into a container"},
{"action": "dunk OBJ into OBJ", "description": "dunk a container into a liquid"},
{"action": "mix OBJ", "description": "chemically mix a container"},
{"action": "go to LOC", "description": "move to a new location"},
{"action": "eat OBJ", "description": "eat a food"},
{"action": "flush OBJ", "description": "flush a toilet"},
{"action": "focus on OBJ", "description": "signal intent on a task object"},
{"action": "wait", "description": "take no action for 10 iterations"},
{"action": "wait1", "description": "take no action for 1 iteration"},
{"action": "inventory", "description": "list your inventory"}
]

Current available actions:
{available_actions}

Tips:
1. You can use look around to see the surroundings of the current location.
2. All containers in the environment have already been opened, you can directly get items from the containers.
3. You can explore the environment and find the items you need to complete the experiment.
**4. Your thinking process should be concise, do not waste time on irrelevant details.**

The target state is: {target_state}

Now it's your turn to take an action. You should first reason step-by-step about the current situation(Your thinking process should be concise and focus on the key information). After you've finished your reasoning, you should choose an appropriate action for the current step and present it within <action> </action> tags.
\end{lstlisting}
\end{tcolorbox}

\section{MobileWorld Experiment}
\label{sec:appendix-mobileworld}

This section provides additional details for the MobileWorld experiment reported in Table~\ref{tab:mobileworld}.

\subsection{Environment and Evaluation}

MobileWorld~\cite{DBLP:journals/corr/abs-2512-19432} is a mobile GUI benchmark that requires agents to interact with Android applications through visual observations and executable UI actions.
Compared with the text-only environments used in our main experiments, MobileWorld introduces richer visual observations, cross-application interaction patterns, and longer-horizon GUI navigation.
For our evaluation, we use the GUI-only subset, comprising 117 instances which focus exclusively on GUI operation and omit user interaction and MCP tasks.

We use MAI-UI-8B~\cite{DBLP:journals/corr/abs-2512-19432} as the base model due to its competent GUI operation capability among open-source models, which provides a good foundation for conducting \textsc{State2State} training.
Given that MobileWorld contains only evaluation tasks and lacks a separate training set, while constructing additional human-annotated training tasks with reliable verifiers would introduce substantial manual costs, we restrict our methodology to \textsc{State2State} training in this environment rather than performing downstream RL on human-specified tasks.
This setup effectively positions MobileWorld as a stress test to evaluate whether \textsc{State2State} can yield robust environment-learning signals without relying on human-authored training tasks.

\subsection{\textsc{State2State} Data}

We adopt a rule-based exploration strategy within MobileWorld. At each step, the strategy randomly selects a clickable XML element, effectively executing a randomized depth-first search (DFS) through the application states.
Each collected state comprises a target screenshot and its corresponding UI hierarchy XML, which jointly serve as the reward signal for \textsc{State2State} training. In total, we collected 365 target states across 14 applications. These were partitioned into a training set of 313 states and a validation set of 52 states, providing a robust dataset for learning from the MobileWorld environment.

\subsection{Reward and Training Setup}

During the training phase, the model receives a target screenshot within its task instruction. At each subsequent interaction step, it operates purely on visual inputs, observing only the current screen image. The underlying UI hierarchy XML remains strictly hidden from the policy and is reserved solely for reward computation.
In our training, both target screenshots and online screen observations are resized to \(540 \times 1200\) to keep the visual context length manageable during multi-turn rollouts.
In evaluation, we use the original image size for standard evaluation settings.

To determine whether the final GUI state matches the target state, the \textsc{State2State} reward combines structural and visual similarity signals. For the structural component, we normalize the XML by removing position- and instance-specific attributes, converting it into a compact semantic sequence composed of resource identifiers, text, content descriptions, and boolean states (e.g., selected or checked). We then calculate a fuzzy sequence similarity score between the current and target XML sequences. For the visual component, we directly compute SSIM-based similarity score between two screenshots. The final reward is computed as a weighted combination, heavily favoring the structural signal with an XML similarity weight of 0.8.

We optimize the policy with GRPO, sampling 6 rollouts per target state with a base batch size of 4 target states. This process is distributed across 24 parallel rollout executions due to computational resource constraints. To enhance training stability, we accumulate gradients over two steps, yielding an effective training batch size of 8 target states. Additionally, we cap the maximum rollout length at 15 interaction steps to prevent excessive sequence lengths and mitigate memory overhead during training.
We set the actor learning rate to \(10^{-6}\), the KL loss coefficient to 0.001, and rollout temperature to 1.0 with top-\(p=0.9\).
The reported \textsc{State2State} model checkpoint is trained for 80 update steps and evaluated on downstream tasks directly, without an additional MobileWorld human-task RL stage.

\subsection{Representative Target States}

\begin{figure*}[t]
  \centering
  \begin{minipage}{0.22\linewidth}
    \centering
    \includegraphics[width=0.9\linewidth]{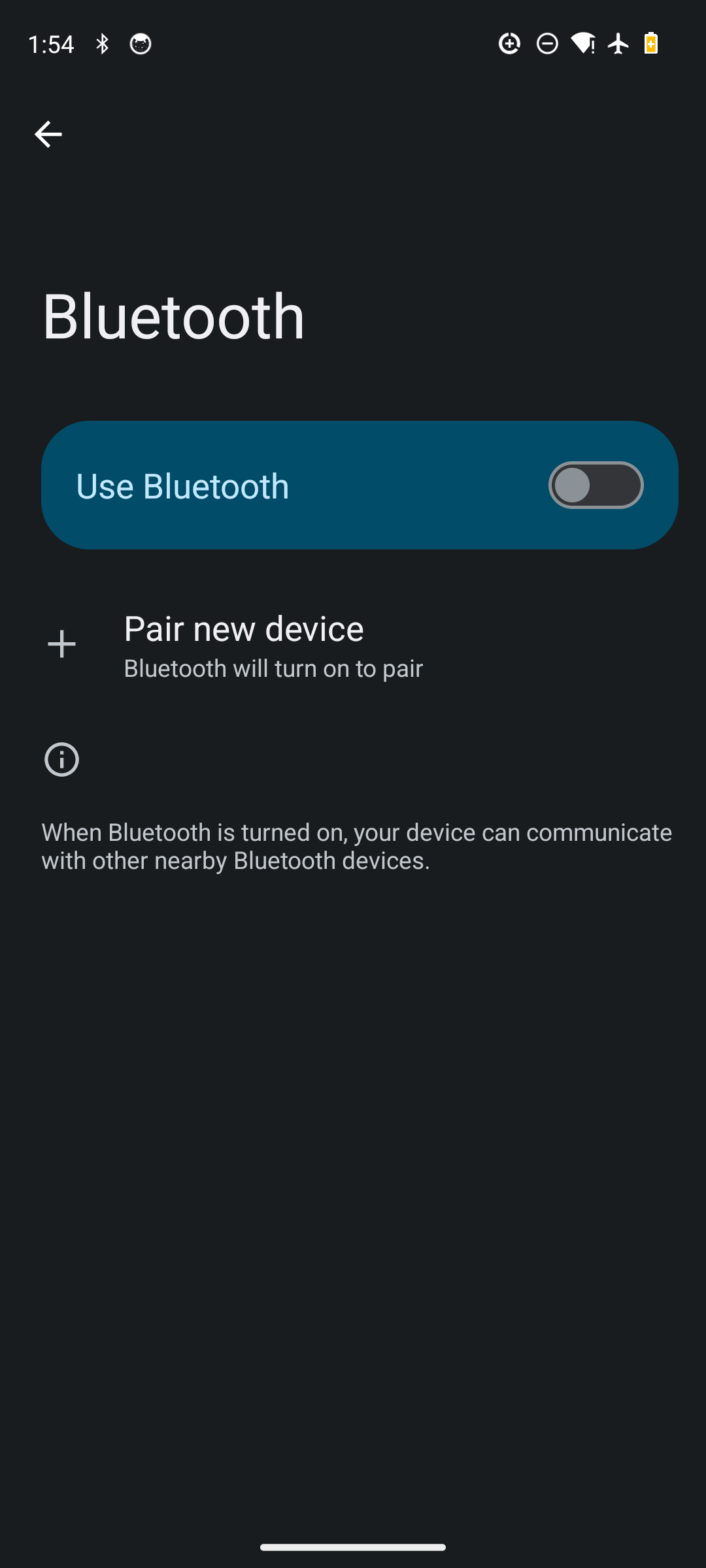}
    \vspace{2mm}
    \small (a) Bluetooth settings.
  \end{minipage}
  \hspace{0.04\linewidth}
  \begin{minipage}{0.22\linewidth}
    \centering
    \includegraphics[width=0.9\linewidth]{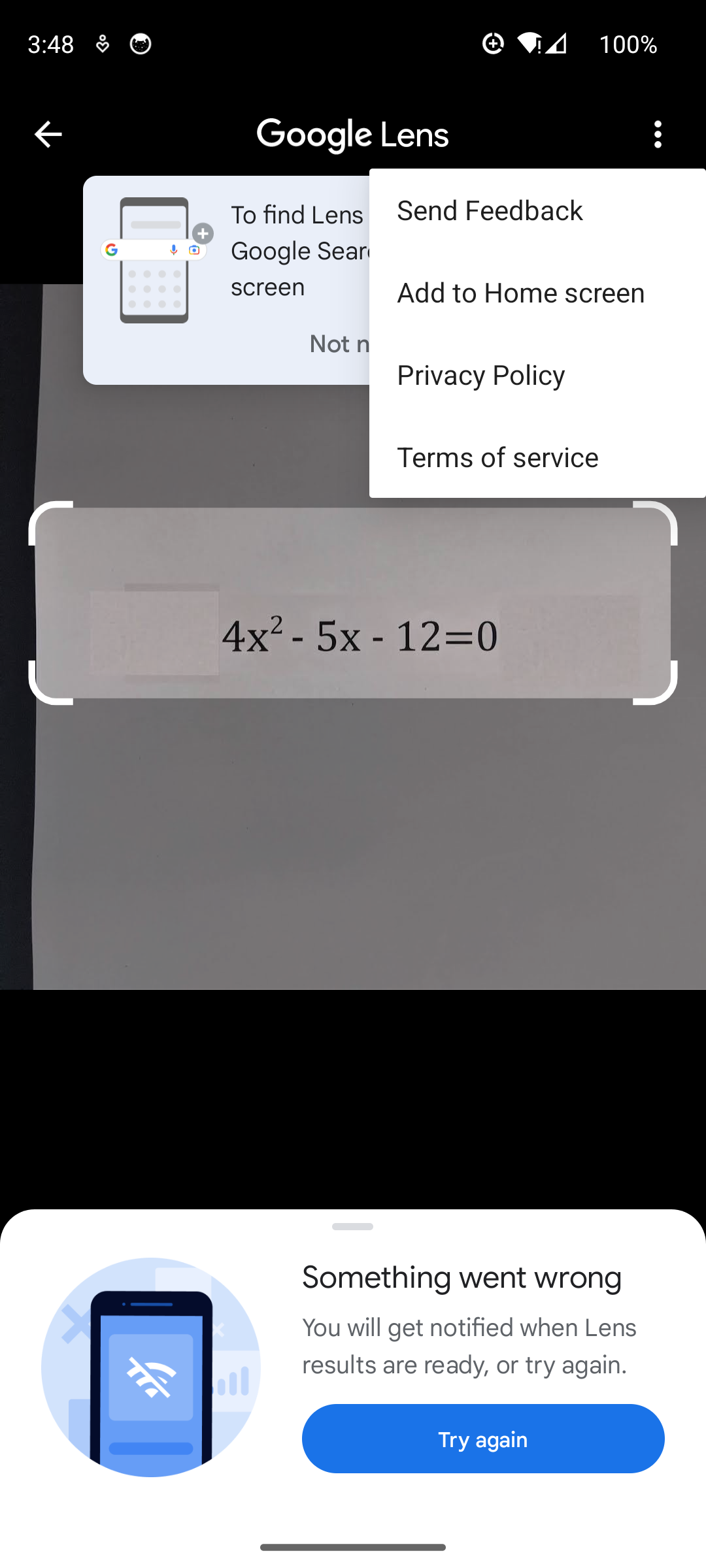}
    \vspace{2mm}
    \small (b) Google Lens.
  \end{minipage}
  \par
  \vspace{3mm}
  \begin{minipage}{0.22\linewidth}
    \centering
    \includegraphics[width=0.9\linewidth]{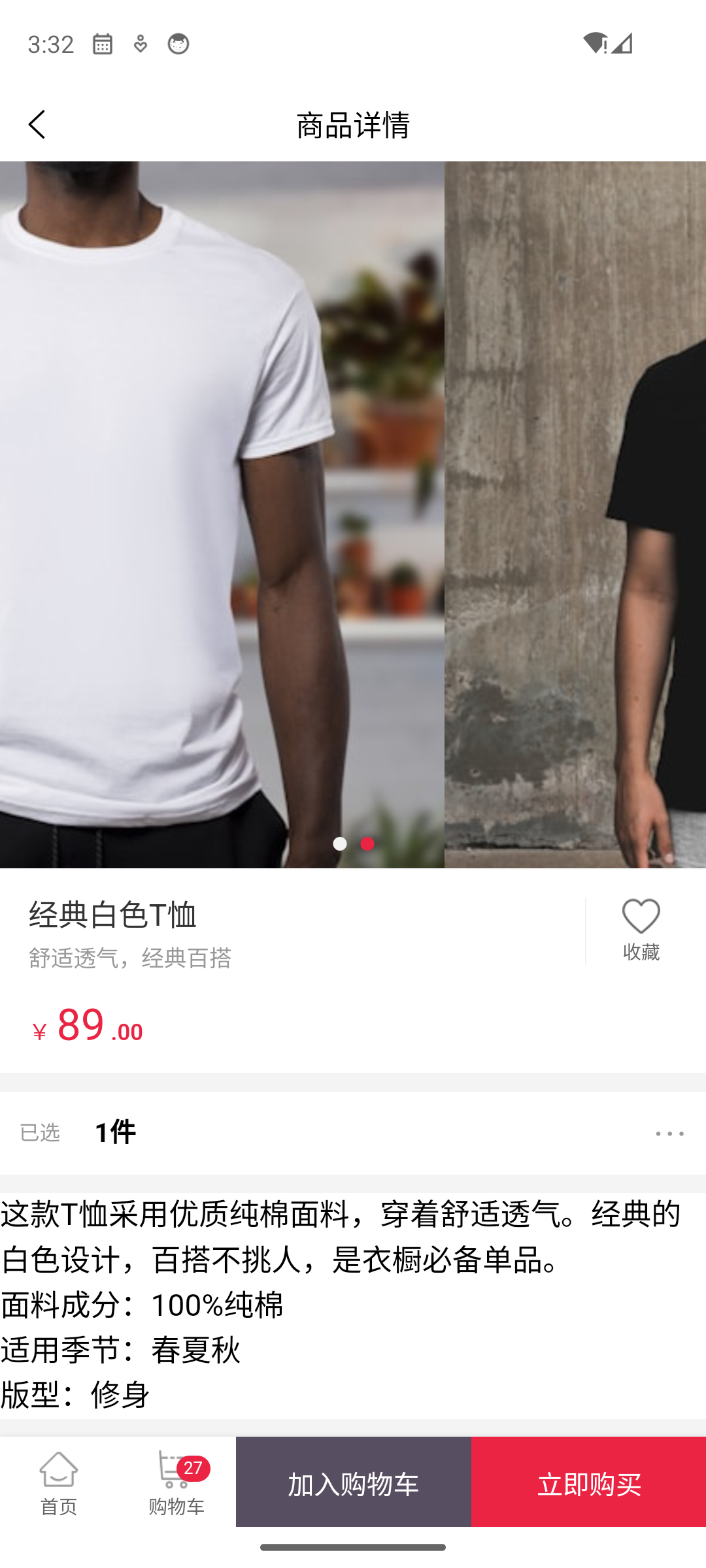}
    \vspace{2mm}
    \small (c) Shopping product page.
  \end{minipage}
  \hspace{0.04\linewidth}
  \begin{minipage}{0.22\linewidth}
    \centering
    \includegraphics[width=0.9\linewidth]{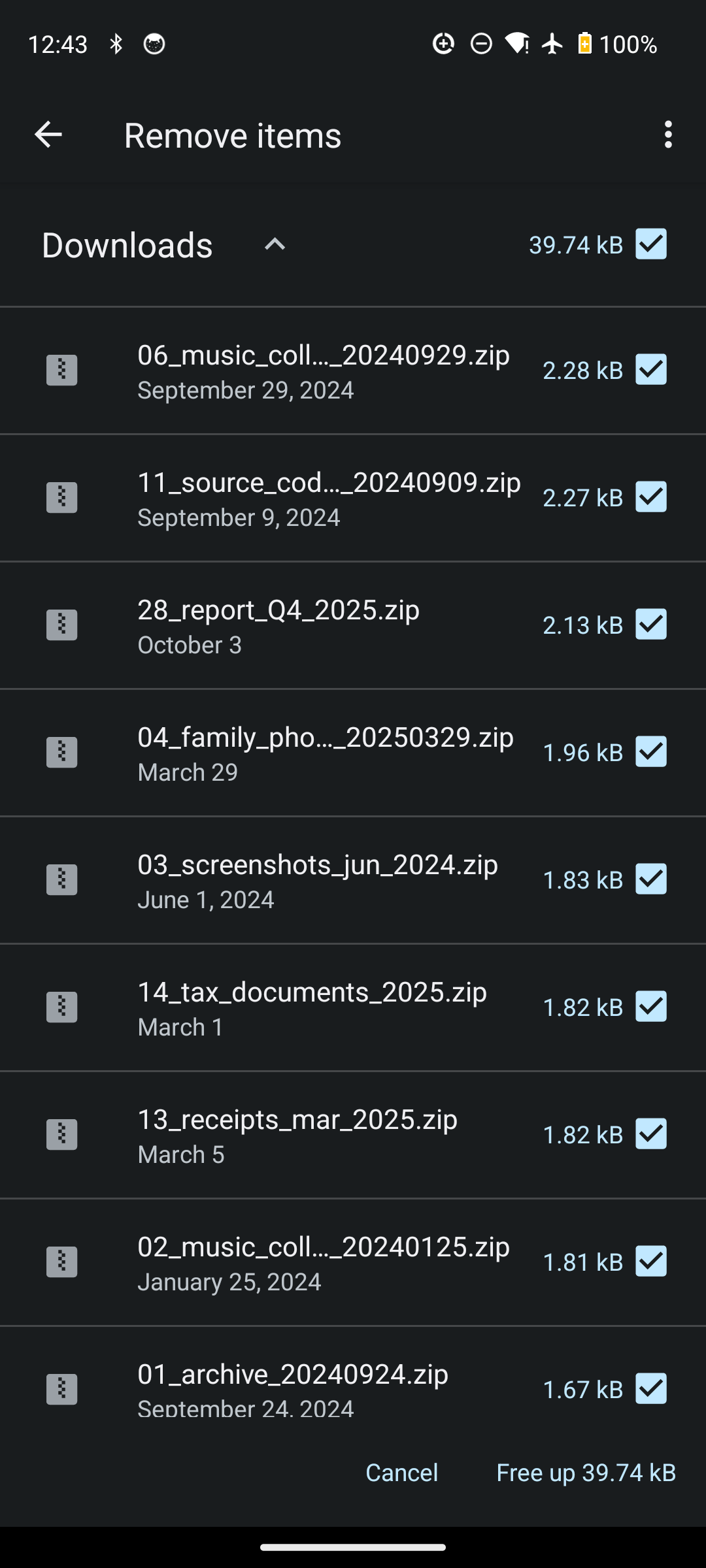}
    \vspace{2mm}
    \small (d) File cleanup selection.
  \end{minipage}
  \caption{Representative MobileWorld \textsc{State2State} target states used for environment-derived training. The model observes a target screenshot and learns to reach a matching GUI state through multi-turn interaction. The examples cover system settings, visual search, shopping, and file-management interfaces.}
  \label{fig:mobileworld-cases}
\end{figure*}

Figure~\ref{fig:mobileworld-cases} shows four representative MobileWorld \textsc{State2State} targets.
These examples illustrate the type of state-reaching objective used during training: the agent is not given a human-written task instruction, but must interact with the phone until the current screen matches the target state.
The targets include system settings, camera-based search, shopping, and file-management states, covering both built-in Android interfaces and application-specific UI modes.
Together, they show that the environment-derived targets span diverse visual layouts, interaction contexts, and application domains.

\section{The Use of Large Language Models (LLMs)}

Large Language Models (LLMs) were used only for language polishing, including grammar checking and improving word- and sentence-level readability. 
All substantive aspects of the paper, including the research ideas, method design, experiments, analysis, and conclusions, were developed and conducted by the authors.

\end{document}